\documentclass[journal]{IEEEtran}

\usepackage{graphicx}
\graphicspath{{Images/}}
\DeclareGraphicsExtensions{.pdf,.jpg,.png}

\usepackage{amsmath,amssymb,amsfonts}

\usepackage[caption=false,font=footnotesize]{subfig}

\usepackage{url}
\usepackage{colortbl}
\definecolor{lr}{rgb}{1, 0.9, 0.9}
\definecolor{lg}{rgb}{0.9, 1, 0.9}

\newcommand{\ignore}[1]{}

\begin{document}

\title{A Deep Learning Framework for Single-Sided Sound Speed Inversion in Medical Ultrasound}

\author{Micha~Feigin, Daniel~Freedman, and Brian~W.~Anthony
\thanks{This work has been submitted to the IEEE for possible publication. Copyright may be transferred without notice, after which this version may no longer be accessible.}%
\thanks{M. Feigin and B. W. Anthony are with the Department
of Mechanical Engineering, Massachusetts Institute of Technology, Cambridge,
MA, USA.}
\thanks{D. Freedman is with Google Research, Haifa, Israel.}}

\maketitle

\begin{abstract}
Objective: Ultrasound elastography is gaining traction as an accessible and useful diagnostic tool for such things as cancer detection and differentiation and thyroid disease diagnostics. Unfortunately, state of the art shear wave imaging techniques, essential to promote this goal, are limited to high-end ultrasound hardware due to high power requirements; are extremely sensitive to patient and sonographer motion, and generally, suffer from low frame rates.

Motivated by research and theory showing that longitudinal wave sound speed carries similar diagnostic abilities to shear wave imaging, we present an alternative approach using single sided pressure-wave sound speed measurements from channel data.

Methods: In this paper, we present a single-sided sound speed inversion solution using a fully convolutional deep neural network. We use simulations for training, allowing the generation of limitless ground truth data.

Results: We show that it is possible to invert for longitudinal sound speed in soft tissue at high frame rates. We validate the method on simulated data. We present  highly encouraging results on limited real data.

Conclusion: Sound speed inversion on channel data has significant potential, made possible in real time with deep learning technologies. 

Significance: Specialized shear wave ultrasound systems remain inaccessible in many locations. longitudinal sound speed and deep learning technologies enable an alternative approach to diagnosis based on tissue elasticity. High frame rates are possible.
\end{abstract}


%


\section{Introduction}
\label{sec:ntroduction}

Mechanical tissue properties, tissue structures, and the spatial arrangement of properties and structures are useful in disease diagnosis in various organs, including the kidneys \cite{linAssociationRenalElasticity2017,singhRenalCorticalElastography2017}, thyroid, muscle, breast \cite{carlsen_strain_2013,changComparisonShearWaveStrain2013}, liver \cite{barrElastographyAssessmentLiver2015,ferraioli_shear_2014}, and prostate. Tracking changes in tissue properties, tissue structure, and the spatial distribution of both is useful for monitoring disease progression as well as response to therapeutic interventions.

As a clinical imaging modality, ultrasound is different from modalities such as CT and MRI in that it uses non-ionizing radiation, is mobile, and has significantly lower purchase and operating costs than most other medical imaging alternatives. The mode of operation is also quite different, as an interactive exploratory approach is taken. The operator can move the probe around, vary applied pressure, and adapt to findings in real time, making real-time quantitative diagnosis techniques that much more important. On the downside, different tissue types are not easily differentiated in the images, requiring more experience to interpret the images.

Embedded in ultrasound signals is information about the mechanical, and acoustic, properties of the tissue through which the ultrasound waves have propagated or from which ultrasound waves have been reflected. Properties include the longitudinal-wave speed of sound, shear-wave speed of sound, tissue density, attenuation, shear modulus, and bulk modulus. As part of the classical B-mode imaging process however, significant parts of this information are discarded through the application of beamforming (delay and sum focusing) and envelope detection.

\begin{figure}
    \center
    \noindent
    \subfloat[\label{fig:goal_a}Channel data input]{
        \includegraphics[width=0.47\columnwidth]{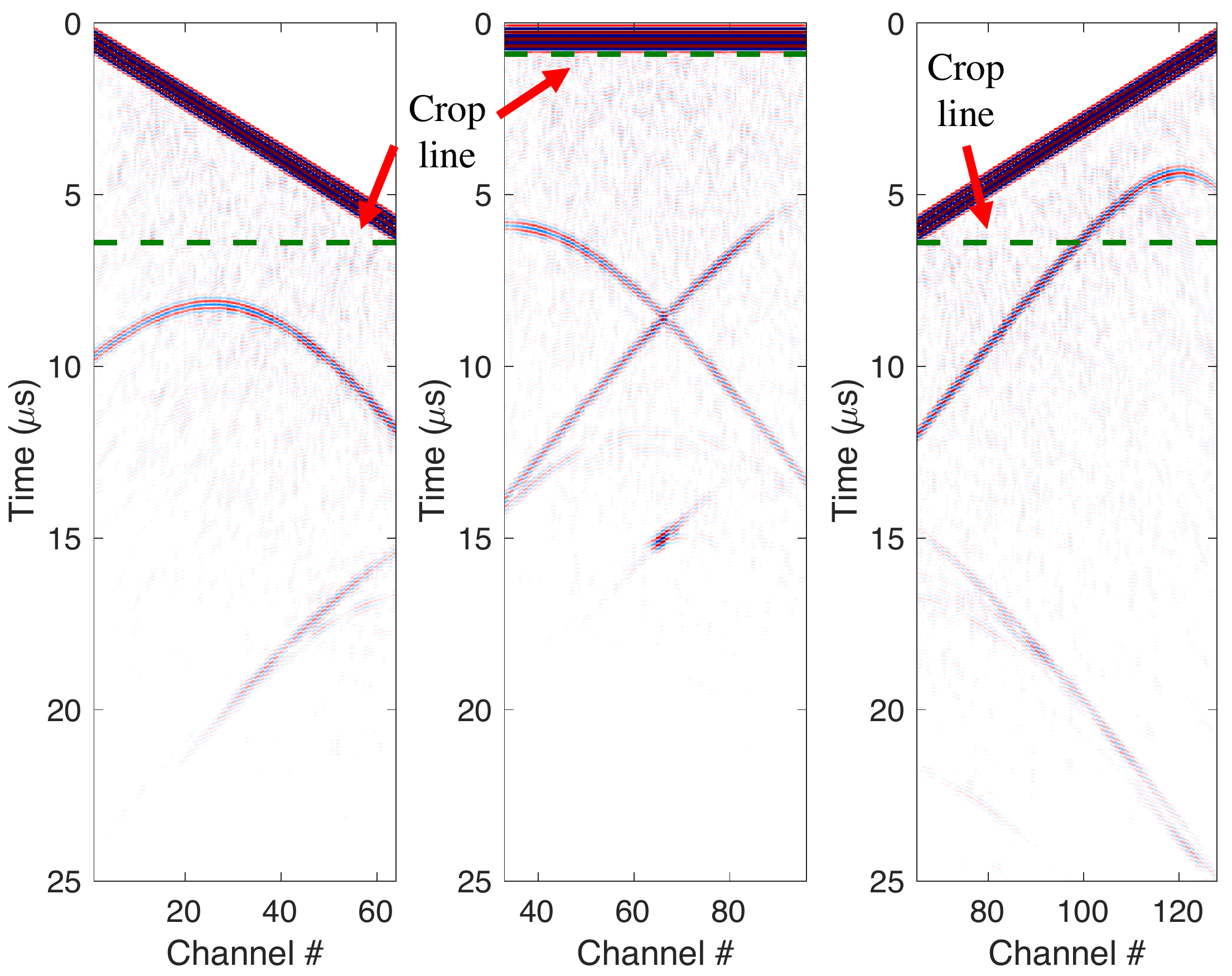}
    }
    
    \subfloat[B-mode image]{
        \includegraphics[height=0.38\columnwidth]{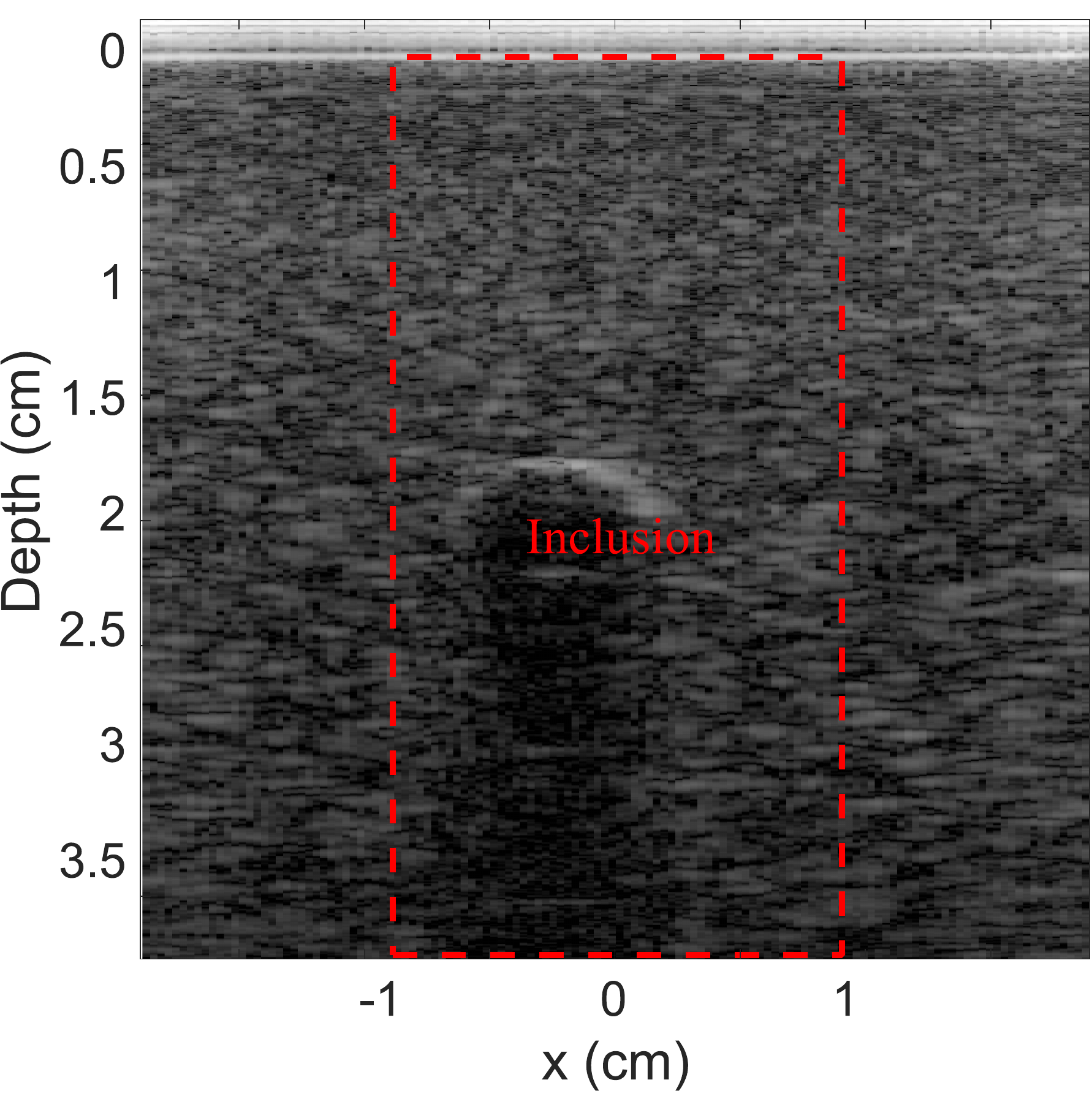}
    }
    \subfloat[Sound speed output]{
        \includegraphics[height=0.38\columnwidth]{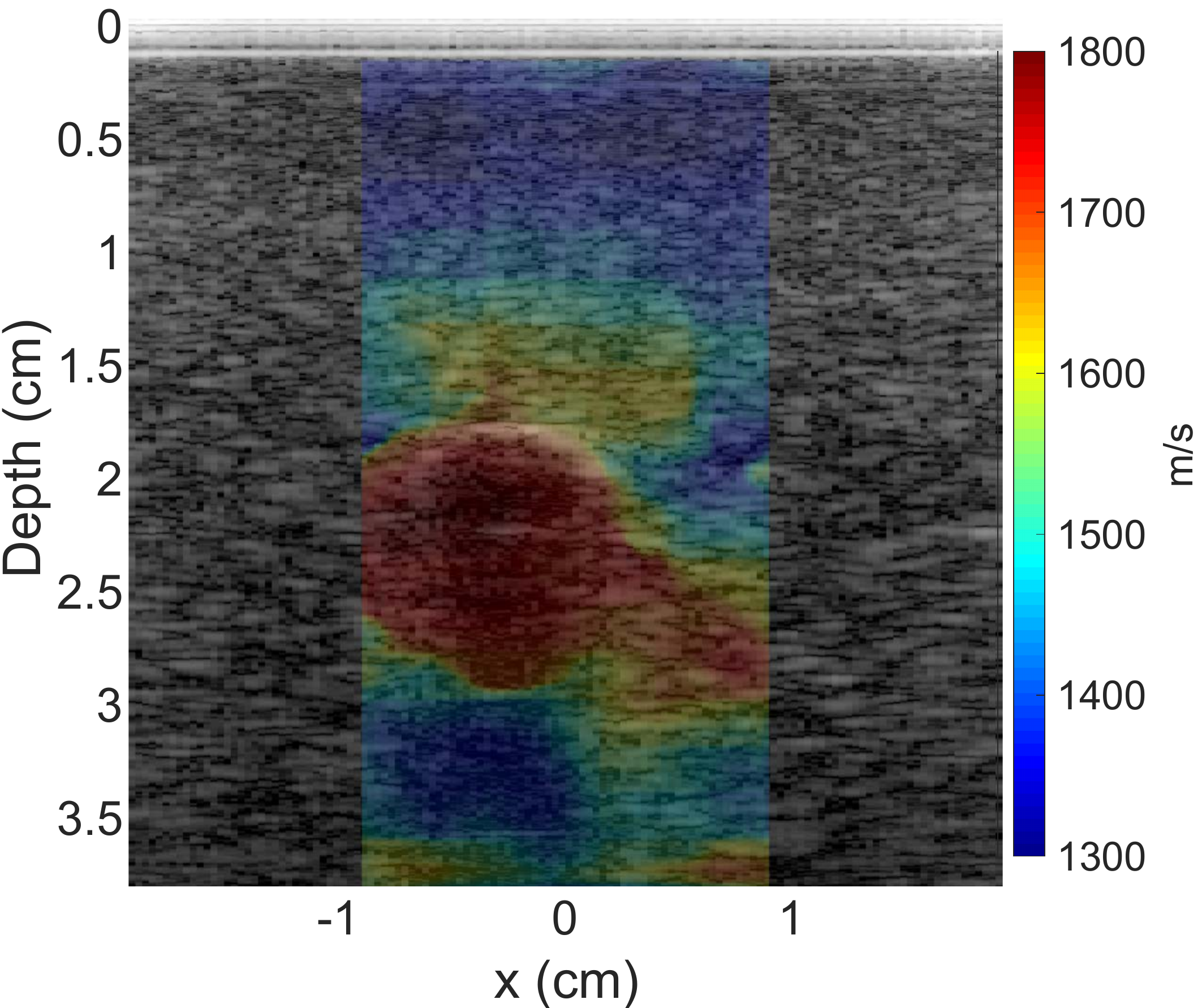}
    }
    
    \caption{\label{fig:goal}Goal: the target of this work is to be able to take raw ultrasound channel data (a) and in addition to the standard B-mode image (b), also produce the corresponding tissue sound speed map (c).}
\end{figure}

In this work, we exploit the information embedded in the raw ultrasound channel data signal. As depicted in Fig.~\ref{fig:goal}, we take ultrasound channel data and generate the sound speed map of the tissue, without going through an explicit imaging or beam-forming step. This resulting information is useful both directly for diagnostic purposes, as well as indirectly, for improving the image formation process, final image quality, and correct for refraction errors. our approach relies on the power of deep convolutional neural networks. We present validation results on simulation data as well as encouraging initial results on real data.

The past several years have seen a burst of interest in the use of artificial inteligence (AI) for improving the physician's workflow, from automatic analysis and improvement of medical images to the incorporation of medical data and physician notes into diagnostics. Considerable research has gone into analyzing image attributes for disease biomarkers. However, the majority of this research effort has taken the imaging process itself as a given and has focused on processing the images coming from a fixed process.  There has been considerably less work on the use of Deep Learning for the direct analysis and processing of raw signals. In ultrasound, the raw signals are the waveforms coming from individual transducer elements, called channel data.

This work is a first step towards learning a full waveform solver for recovering  elastic and viscoelastic tissue parameters using Deep Learning and shows the viability of this approach.  Despite numerous applications for various inverse problems within different image domains, this is the first work we are aware of which applies a Deep Learning framework to the analysis of raw time domain RF type signals.

An advantage of our approach is that it can work at real-time frame rates, with the current implementation running at over 150fps on a single NVIDIA 1080Ti GPU. It requires only a small number of transmits, and can, therefore, run in parallel to standard imaging. The physical limitation on frame rates is a function of tissue depth, but is thousands of  frames per second. This opens the door to dynamic functional imaging.

\section{Background and prior work}
\label{sec:motivation}

\subsection{General Background}

Our goal in this work is to measure physical tissue properties with diagnostic relevance. The currently deployed approach in medical ultrasound is shear wave elastography \cite{greenleaf_selected_2003,kathynightingaleAcousticRadiationForce2011,gennisson_ultrasound_2013}. Shear wave imaging is based on the measuring the speed at which the shear wave front propagates in tissue. The shear wave speed is directly dependent on the shear modulus; it is used to approximate Young's modulus by employing several assumptions, mainly, isotropy, incompressibility, and a known density. Young's modulus is the value most closely related to what we intuitively perceive as material stiffness, and is thus related to the physician's notion of palpation based diagnostics.

An alternative approach, used mostly in the field of breast imaging, is travel time tomography \cite{duric_breast_2013,sakUsingSpeedSound2017}. Here, the speed of compression, or longitudinal, waves in tissue is measured based on the acoustic travel times between known location pairs. Longitudinal speed of sound measurements can be viewed, depending on the case at hand, as either an alternative or a complementary approach to shear wave imaging, as the longitudinal sound speed is also related to Young's modulus. Among other things, variations of longitudinal sound speed in fat are more distinct than shear sound speed, making it relevant for diagnosing liver and kidney diseases, such as non-alcoholic fatty liver disease (NAFLD), as well as degenerative muscle diseases such as Duchenne's muscular dystrophy.

In this work, we take the approach of recovering speed of sound information. We however bypass tomographic imaging and take the path of  single sided sound speed recovery, similar to seismic imaging \cite{yilmaz_seismic}. We use numerical simulations to train a deep learning network to extract the speed of sound information present in the raw reflected pressure waves. Some examples of information regarding sound speed, independent of comparing geometric travel distance to travel times, includes the geometry and wave front deformation due to sound speed and refraction, the presence and appearance of head waves (waves traveling along the interface with a higher velocity medium), critical reflection angle, reflection amplitude and sign (positive or negative reflection) as well as the reflection coefficient variation as a function of the angle of incidence, called amplitude versus angle (AVA) or amplitude versus offset (AVO). Probably the easiest of these to understand is the wave front geometry of a point scaterrer. These appear as hyperbolas in the space/time plot, similar to those seen in Fig.~\ref{fig:goal_a}. The angle between the asymptotes depends only on the sound speed, while the structure of the apex is also dependent on the distance from the probe to the scatterer. Refracted waves deform the wavefront, providing both a source of information on sound speed, as well as motivation for being able to correct for sound speed variations.

Next, we highlight specific aspects of prior work.  We begin by with an overview of the physics and techniques for shear wave elastography and tomographic ultrasound elastography. We summarize the clinical motivation for using these metrics. Finally, we give an overview of deep learning in the context of medical imaging.

\subsection{Elastography and Full Waveform Inversion}

The prevailing model for ultrasound elastography of soft tissue is that of a linear isotropic elastic material \cite{greenleaf_selected_2003}. While this model does not account for non-linear effects, it is still useful for diagnostic purposes for many soft tissues.

Under this model, tissue properties can be described using density and two independent elastic coefficients. Some of the commonly used pairs are Young's modulus ($E$) and Poisson ratio ($\nu$), bulk modulus ($K$) and shear modulus ($G$), and the two Lamme parameters ($\lambda$ and $\mu$ - where $\mu$ is the shear modulus). Young's modulus is most often used to describe tissue stiffness. The complementary elastic parameter, either the Poisson ratio or bulk modulus, as well as the density, are often assumed to be constant in soft tissue imaging, and dominated by the tissue's high water content.

The pressure wave (also known as p-wave, primary wave or longitudinal wave) is an acoustic wave used for ultrasound imaging and travels at $1540\,m/s \pm 10\%$ on average in soft tissue. The shear wave (also known as s-wave, secondary wave or transverse wave) is measured indirectly in ultrasound elastography, using pressure waves, and is much slower. It travels at velocities on the order of $3\,m/s$ in healthy tissue, and up to $60\,m/s$ in highly pathological tissue. When working in the linear acoustics regime in soft tissue, as is the case for medical ultrasound imaging, the pressure waves and the two orthogonally polarized shear waves are independent, and only distinctly couple at strong discontinuities. Note however that this independence is only partially correct, as the pressure waves actually ``see'' the displacement caused by the shear waves. This, together with the vast sound speed difference between the two wave types, allows the use of pressure waves to image shear wave propagation and is the basis for shear wave imaging.

Common methods of shear wave generation include acoustic radiation force (ARFI) \cite{kathynightingaleAcousticRadiationForce2011} and supersonic shear wave imaging \cite{bercoff_supersonic_2004,nahas_supersonic_2013}. A mechanical shear wave is generated in the tissue and its propagation speed is tracked using pressure waves. These methods, however, are limited to high-end devices due to high power and probe requirements. They also generally suffer from low frame rates, long settling times, and high sensitivity to sonographer and subject motion.

Tomographic ultrasound imaging for travel time tomography and full waveform inversion (FWI) are related and actively researched techniques. Travel time tomography measures first arrival times between a set of known transmitter-receiver pairs. This travel time depends on the integral on slowness (reciprocal of the sound speed) along the geodesic. FWI performs optimization on a tissue model, to minimize the residual between the measured signal and the simulation and is not dependent on knowing travel distances. This research is currently mostly focused on breast \cite{duric_breast_2013,sakUsingSpeedSound2017,roySoundSpeedEstimation2010,li_multi-grid_2012,hoppBreastTissueCharacterization2012,nebekerImagingSoundSpeed2012,li_toward_2014}, and musculoskeletal \cite{finckeUltrasoundTravelTime2016,finckeImagingCorticalBone2017} imaging, both showing promising prospects. Current in vivo implementations require a full circumferential field of view, limiting them to small body parts. Both are computationally expensive, with FWI being more costly. FWI is sensitive to noise, choice of initial conditions and has a harder time with piecewise constant velocities, while travel time tomography is sensitive to lensing effects. One example is limb imaging, where the first arrival travels in the bone layer around the marrow, shadowing the signal traveling through the marrow, making it difficult to impossible to image the bone marrow.

Single-sided techniques are in use in the seismic domain, but often require operator intervention for good results. Their use in medical ultrasound imaging is limited, and is generally performed based on focusing techniques or PSF analysis on the final b-mode image, and not by an inversion method \cite{andersonDirectEstimationSound1998,hachiyaDeterminationSoundSpeed1992,shinEstimationSpeedSound2010,alexbenjaminSurgeryObesityRelated2018,zubajloSurgeryObesityRelated2018}. One of the few exceptions is the CUTE method that uses the wavefront deformation as seen in the reflection pattern of scatterers \cite{Preisser2014,Stahli2019}.

The motivation for these approaches can be understood by looking at the dependence of the shear wave and longitudinal wave sound speeds on the underlying physical properties:
\begin{align}
C_{\mbox{longitudinal}} &= \sqrt{\frac{K + \frac{4}{3} G}{\rho}} \\
C_{\mbox{shear}} &= \sqrt{\frac{G}{\rho}}
\end{align}
where $C$ denotes the appropriate speed of sound. Under the previously stated assumption of a constant bulk modulus and density, at least to a first order approximation, both squared velocities depend linearly on the same single value, the shear modulus, which is in turn directly related to the value of interest, Young's modulus.

\subsection{Clinical Motivation}

Short of attaining fully automated diagnostics, the next best thing is to solve the inverse problem of measuring physical tissue properties. Focus is given to properties that can be used directly by the physician as reliable disease biomarkers. Achieving that end in an accessible and easily undertaken way can greatly improve the physician workflow as well as make quality health care much more accessible.

Again, current research is split in two main directions, shear wave elastography and ultrasound tomography, both travel time tomography and full waveform inversion. Table~\ref{table:youngs_modulus} presents Young's modulus values for several healthy and pathological types of breast tissue as given by \cite{samani_elastic_2007,fung_biomechanics_1993}. Young's modulus has a strong predictive value for detecting and differentiating pathological tissue.

\begin{table}
\caption{\label{table:youngs_modulus}Young's modulus as a disease biomarker for various breast tissue types \cite{samani_elastic_2007,fung_biomechanics_1993} }

\centering %
\noindent %
\begin{tabular}{|c|c|c|}
\hline 
Brest tissue type  & \# of samples  & Young's modulus (kPa) \tabularnewline
 &  & mean $\pm$ STD\tabularnewline
\hline 
\hline 
Normal fat  & 71  & 3.25 $\pm$ 0.91 \tabularnewline
Normal fibroglandular tissue  & 26  & 3.24 $\pm$ 0.61 \tabularnewline
Fibroadenoma  & 16  & 6.41 $\pm$ 2.86 \tabularnewline
Low-grade IDC  & 12  & 10.40 $\pm$ 2.60 \tabularnewline
ILC  & 4  & 15.62 $\pm$ 2.64 \tabularnewline
DCIS  & 4  & 16.38 $\pm$ 1.55 \tabularnewline
Fibrocystic disease  & 4  & 17.11 $\pm$ 7.35 \tabularnewline
Intermediate-grade IDC  & 21  & 19.99 $\pm$ 4.2 \tabularnewline
High-grade IDC  & 9  & 42.52 $\pm$ 12.47 \tabularnewline
IMC  & 1  & 20.21\tabularnewline
Fat necrosis  & 1  & 4.45\tabularnewline
\hline 
\end{tabular}
\end{table}

Researchers have also shown that longitudinal wave sound speed has similar diagnostic ability to shear wave imaging \cite{sakUsingSpeedSound2017,hachiyaRelationshipSpeedSound1994,matsuhashiEvaluationHepaticUltrasound1996,li_vivo_2009,imbaultRobustSoundSpeed2017,benjaminNoninvasiveDiagnosisNonalcoholic2017}. Some of these results taken on breast tissue are presented in Table~\ref{table:sound_speed} (from \cite{li_vivo_2009}).

\begin{table}
\caption{\label{table:sound_speed}Longitudinal sound speed as a disease biomarker for various breast tissue types \cite{li_vivo_2009} }

\centering %
\noindent %
\begin{tabular}{|c|c|}
\hline 
Brest tissue type  & Sound speed (m/s)\tabularnewline
\hline 
\hline 
Normal fat  & $1442\pm9$ \tabularnewline
Breast parenchyma & $1487\pm21$\tabularnewline
Benign breast lesions & $1513\pm27$\tabularnewline
Malignant breast lesions & $1548\pm17$\tabularnewline
\hline 
\end{tabular}
\end{table}

Using the longitudinal speed of sound as a substitute for the transverse speed of sound presents several potential advantages. Longitudinal waves travel significantly faster in tissue than transverse waves, allowing for much higher frame rates. Transverse waves cannot be detected directly by the probe, due to their strong attenuation in tissue along with low sensitivity of the sensor ellements to transverse motion. As a result, they are only imaged indirectly based on their effect on longitudinal waves. The particle motion that is detected is on the order of 10 microns, on the order of $1/30$ of a wavelength, resulting in measurements that are highly sensitive to probe and subject motion. The amount of energy required to generate shear waves using acoustic radiation force is also high, requiring correspondingly high powered devices.  This, in turn, limits this technology to high-end ultrasound machines; furthermore, frame rates must be lowered due to FDA limitations on transmission power, tissue heating, and long settling times.

\subsection{Deep Learning}
\label{sec:deep_learning}

The astounding success of Deep Learning in fields including computer vision, speech recognition, and natural language processing is by now widely known.  Neural networks have achieved state of the art results on many benchmarks within each of these fields.  In most cases, the networks in question are relatively deep (tens or hundreds of layers) and are trained by stochastic gradient descent or related techniques, as implemented by the standard backpropagation algorithm.

Deep Learning has also achieved great success in medical imaging on standard computer vision tasks, such as classification \cite{wang2017chestx}, detection \cite{byrne2018mo1679}, and segmentation \cite{havaei2017brain}.  However, only recently has Deep Learning been applied to problems in sensing and image reconstruction.  The growing popularity of this trend is exemplified by the recent special issue (June 2018) of TMI which was devoted to this topic.  The issue contained many papers related to both CT and MRI.  Within the realm of CT, a variety of topics were examined, including artifact reduction \cite{zhang2018convolutional}, denoising \cite{yang2018low,kang2018deep}, and sparse-view \cite{zhang2018sparse,han2018framing} and low-dose \cite{zheng2018pwls,shan20183d} reconstruction.  Papers on MRI tended to focus on Deep Learning approaches to compressive sensing \cite{yang2018dagan,gozcu2018learning,quan2018compressed}.  Deep Learning has also been applied, though not quite as widely, to PET \cite{yang2018artificial,kim2018penalized} and Photoacoustic Tomography \cite{hauptmann2018model,allman2018photoacoustic}.  Furthermore, we note that in the broader signal processing community, work has been devoted to applying Deep Learning techniques to general reconstruction problems, such as compressive sensing \cite{mousavi2017deepcodec} and phase retrieval \cite{metzler2018prdeep}.

Within the field of ultrasound, Deep Learning has been successfully employed in a few areas.  Vedula \emph{et al.} \cite{vedulaCTqualityUltrasoundImaging2017} train a multi-resolution CNN to generate CT quality images from raw ultrasound measurements.  Yoon \emph{et al.} \cite{yoonEfficientBmodeUltrasound2017} apply Deep Learning to produce B-mode images from a small number of measurements, which can circumvent the heavy computational load imposed by competing compressed sensing algorithms.  Tom and Sheet \cite{tom2018simulating} propose a method for generating B-mode images based on generative adversarial networks, which obviates the need for running expensive wave equation solvers.  Luchies and Byram \cite{luchiesDeepNeuralNetworks2018} apply Deep Learning to the problem of beamforming in ultrasound, to minimize off-axis scattering.  Reiter and Bell  \cite{reiterMachineLearningApproach2017} use a CNN to identify a common type of ultrasound artifact, namely the reflection artifact that arises when a small point-like target is embedded in the middle of an echogenic structure.

Finally, we note that Deep Learning has also been applied in non-reconstruction tasks to ultrasound, including classification \cite{ravishankar2016understanding,zheng2018transfer} and segmentation \cite{xian2018benchmark}.

\section{Simulations}
\label{sec:simulations}

Collecting real-world ground truth ultrasound data in quantities sufficient for training a neural network, is practically impossible. This leaves us with the alternative of using simulated data.

For our simulations, the sensor is modeled based on our physical system, a Cephasonics cQuest Cicada ultrasound system, capable of transmitting and receiving on 64 channels at a time. The ultrasound probe is a 1D linear array transmitting at a central frequency of $5 MHz$. The probe face is $3.75\,cm$ wide and contains 128 elements. We locate the probe plane in the simulations just outside the perfectly matched layer (PML - used to numerically absorb waves incident on the boundary) with four grid points per Piezo element and an extra four grid points for the kerf (spacing between elements). The total grid dimension is $4.24\,cm$ by $4.24\,cm$ or 1152 by 1152 elements. The simulation is run in 2D.

\begin{figure}
        \center
        \noindent
        \subfloat[Sound speed]{
            \includegraphics[width=0.47\columnwidth]{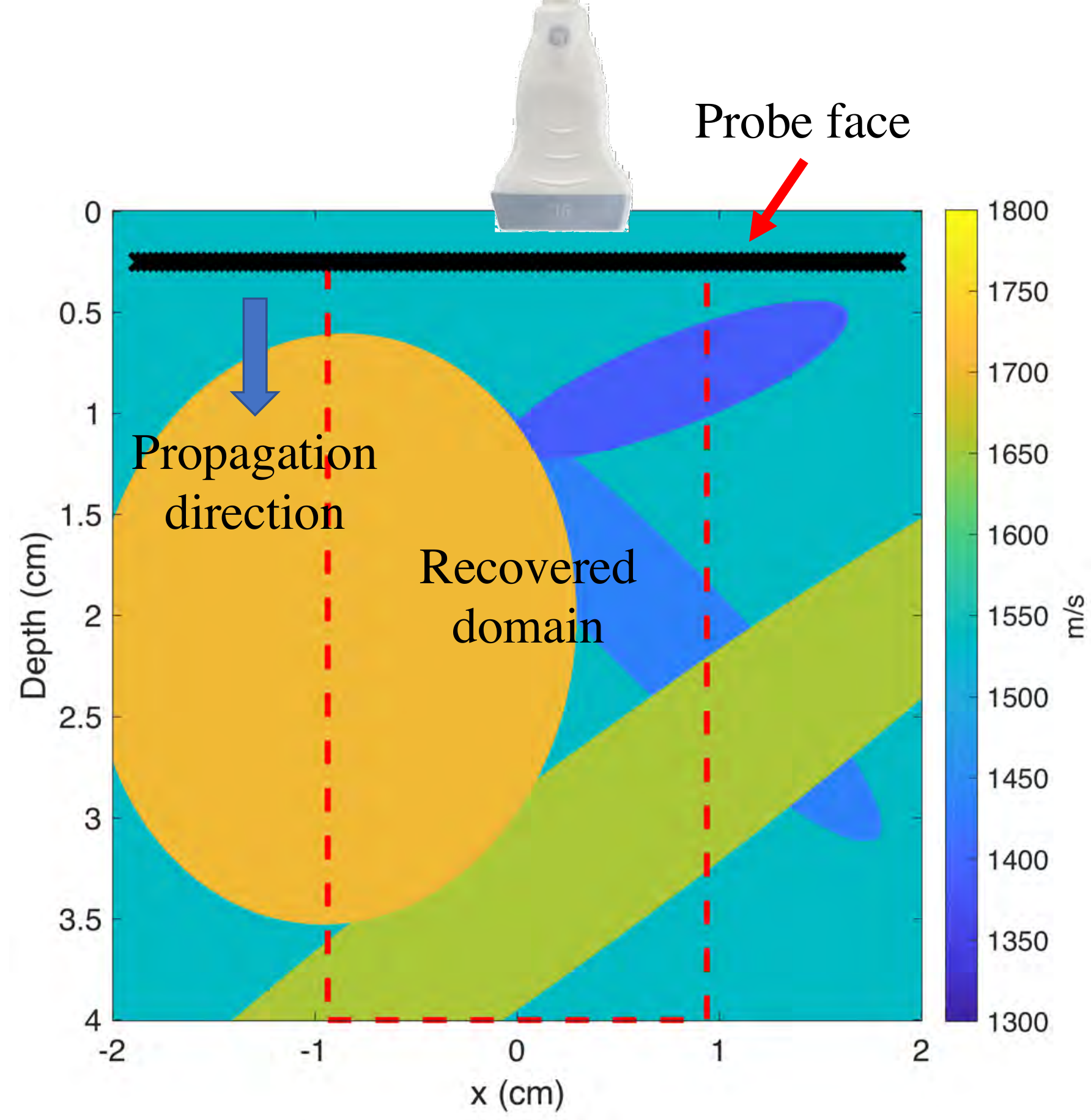}
        }%
        \subfloat[Speckle]{
            \includegraphics[width=0.47\columnwidth]{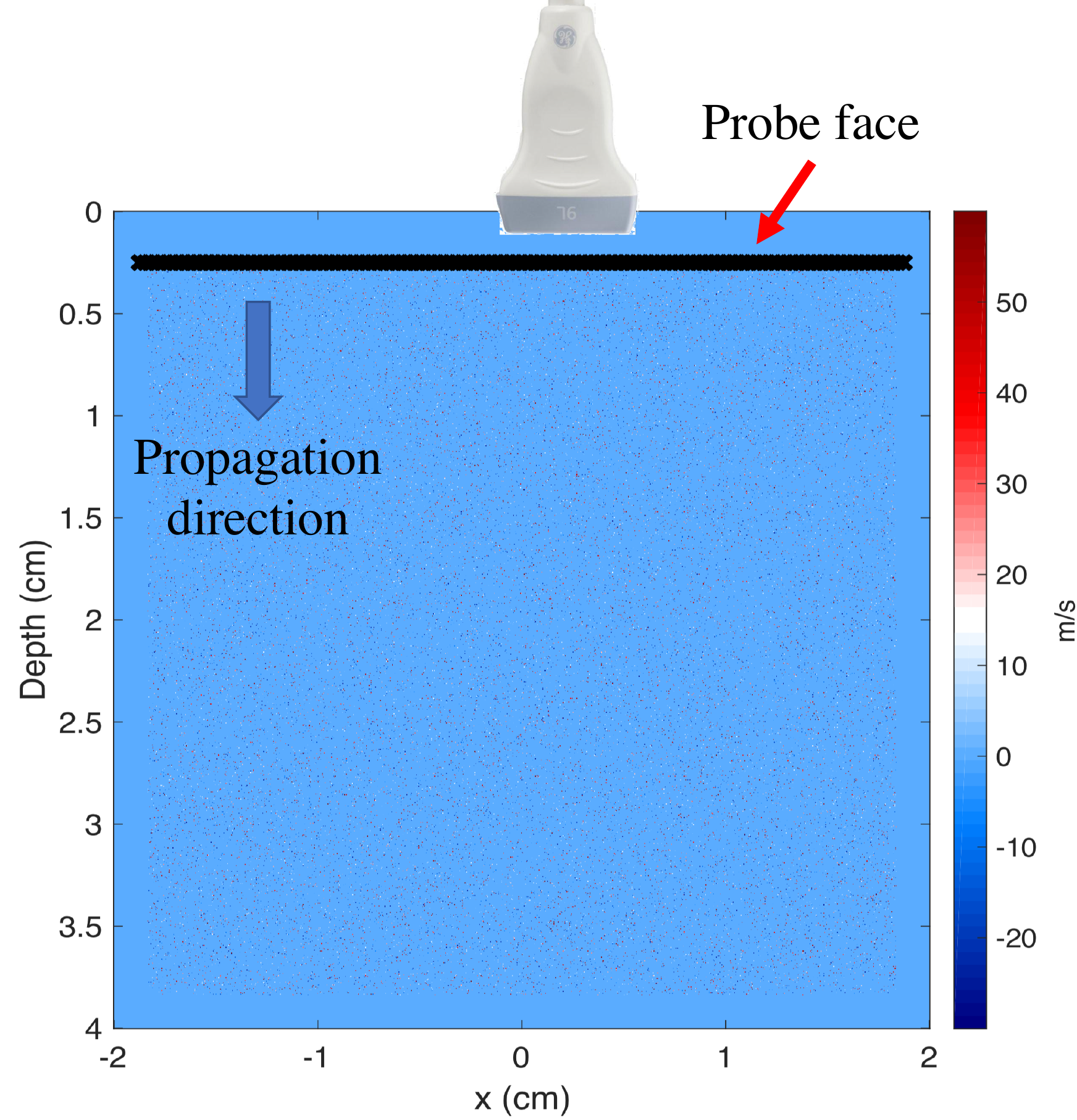}
        }

        \caption{\label{fig:simulation}Simulation setup. Reflecting objects are defined in the sound speed domain (a). Ultrasound speckle is defined in the density domain (b). The probe face is at the top end of the domain, marked by hash marks, and is outside the PML, propagation is towards the bottom. The recovered sound speed domain is marked by a red dashed line in (a).}
\end{figure}

To generate the training data, we use a simplified soft tissue model for organs and lesions. The emphasis is on a model that can generate a large and diverse set of random samples. We model organs in tissue as uniform ellipses over a homogeneous background. The mass density is set to $0.9\,g/cm^3$. Between one and five ellipses (organs) are randomly placed. The sound speed for the background and each of the ellipses is randomly selected based on a uniform distribution in the range of $1300\,m/s$ and $1800\,m/s$. Random speckle is generated in the density domain with a uniformly distributed mass density variations between $-3\%$ and $+6\%$ and a mean distribution density of 2 reflectors per wavelength ($\lambda$) squared. Attenuation is fixed at $0.5\, dB / (MHz \cdot cm)$, or $2.5\,dB / cm$ at the center frequency of $5\,MHz$. For the recovered domain we chose the central section, $1.875\,cm$ wide by $3.75\,cm$ deep, with a $3.75 cm$ wide probe. This is guided by two considerations: (1) coverage limitations due to maximum aperture size; and (2) the desire to show that our method can handle signals arriving from outside the recovered domain. The setup is depicted in Figure~\ref{fig:simulation}.

The numerical solver we work with is the k-wave toolbox for MATLAB \cite{treebyKWaveMATLABToolbox2010,treebyModellingElasticWave2014}. It presents a compromise that can deal with both discontinuities as well as speckle noise over non-uniform domains while maintaining decent run times on an NVIDIA GPU.

\begin{figure}
        \noindent
        \center
        \subfloat[Left plane wave]{
                \includegraphics[viewport=0bp 0bp 220bp 212bp,clip,width=0.3\columnwidth]
                        {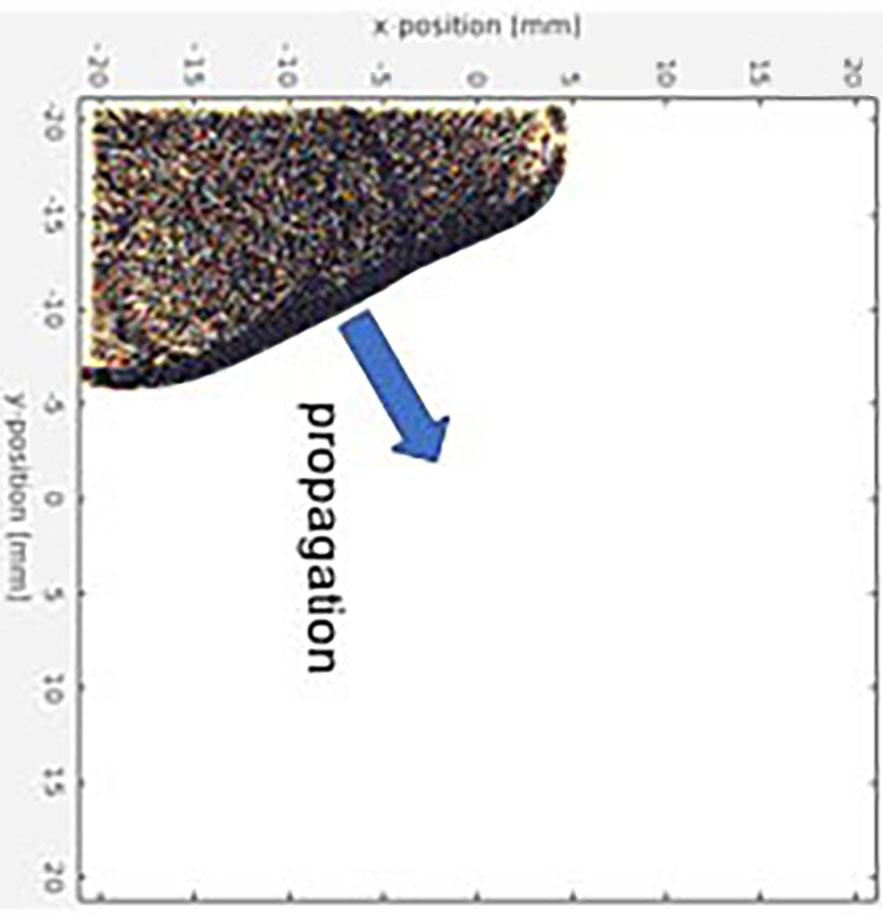}
        }%
        \subfloat[Middle plane wave]{
                \includegraphics[viewport=0bp 0bp 220bp 212bp,clip,width=0.3\columnwidth]
                        {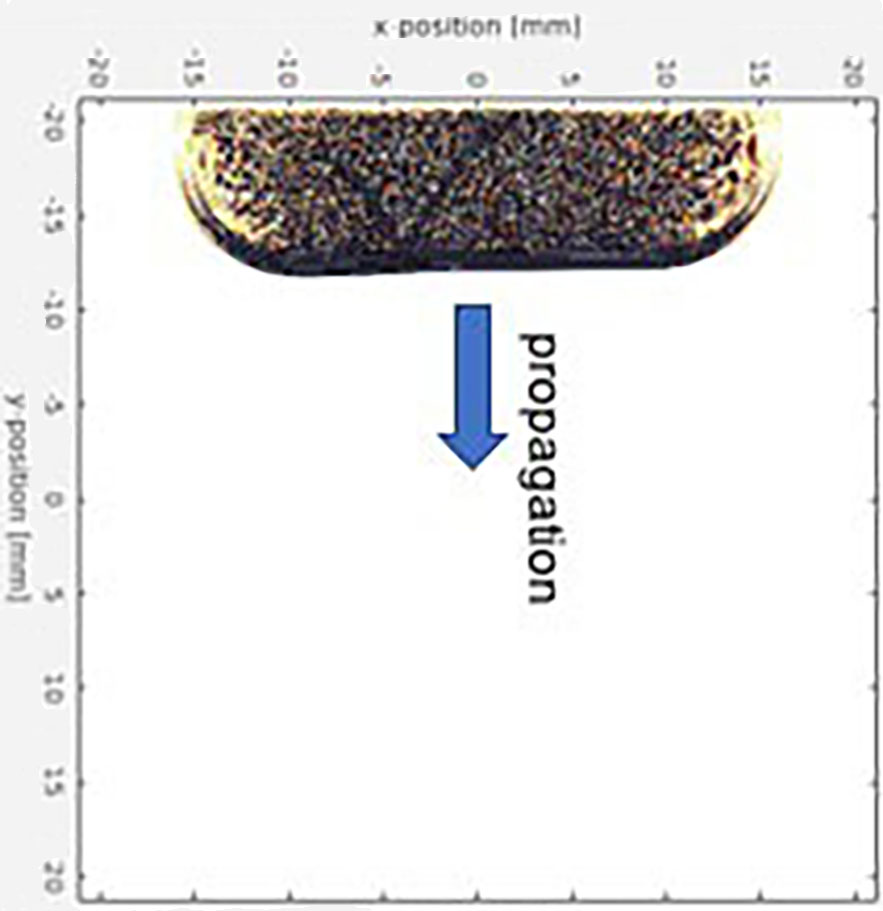}
        }
        \subfloat[Right plane wave]{
                \includegraphics[viewport=0bp 0bp 220bp 212bp,clip,width=0.3\columnwidth]
                        {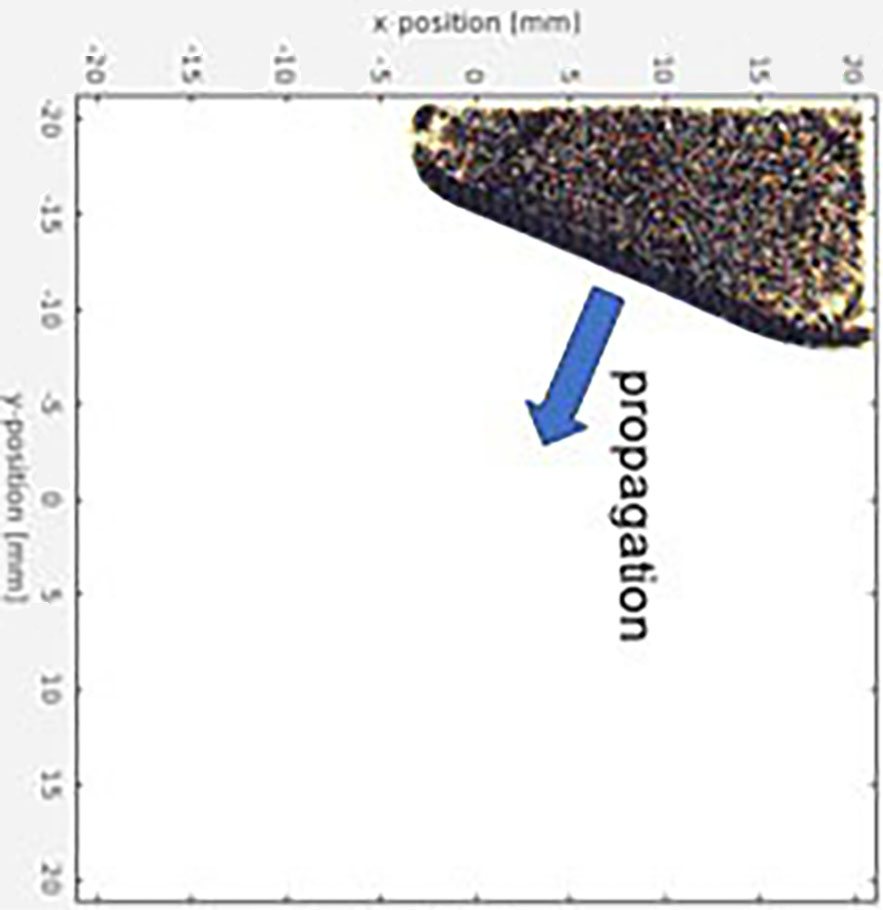}
        }

        \caption{\label{fig:plane_wave}The diagonal three plane waves generated in k-wave as well as by the real probe. Each plane wave is generated by 64 elements (half the probe), the limit of our current system.}
\end{figure}

For the transmit pattern we are limited by three parameters: simulation time, network resources, and signal to noise ratio (SNR). Both simulation time, as well as network run-time, training time, and resources, are controlled by the number of pulses. In this work, we investigate the sound speed recovery quality using either one or three transmit pulses. This prevents us from using the classic scanning focused beam imaging approach. Due to SNR issues, using point source transmits, \emph{i.e.} transmitting from a single element, is also problematic. As a result, we choose to work with three plane waves: one direct plane wave transmitted from the center of the probe, and two diagonal plane waves from the edges. These plane waves are depicted in Figure~\ref{fig:plane_wave}. The plane wave angle is chosen to best cover the full domain.

\section{Network Setup}
\label{sec:network}

We now describe the structure of our neural network.  We wish to map signals to signals, hence we use a type of fully convolutional neural network (FCNN) - that is, there are no fully-connected layers, only convolutional layers, in addition to various non-linear activations.  However, most FCNNs assume the input and output sizes are identical, whereas that is not the case in our setup.  Therefore, we use striding within the convolutions to effectively achieve decimation.

\begin{figure}
    \noindent
    \center
       \includegraphics[scale=0.24]{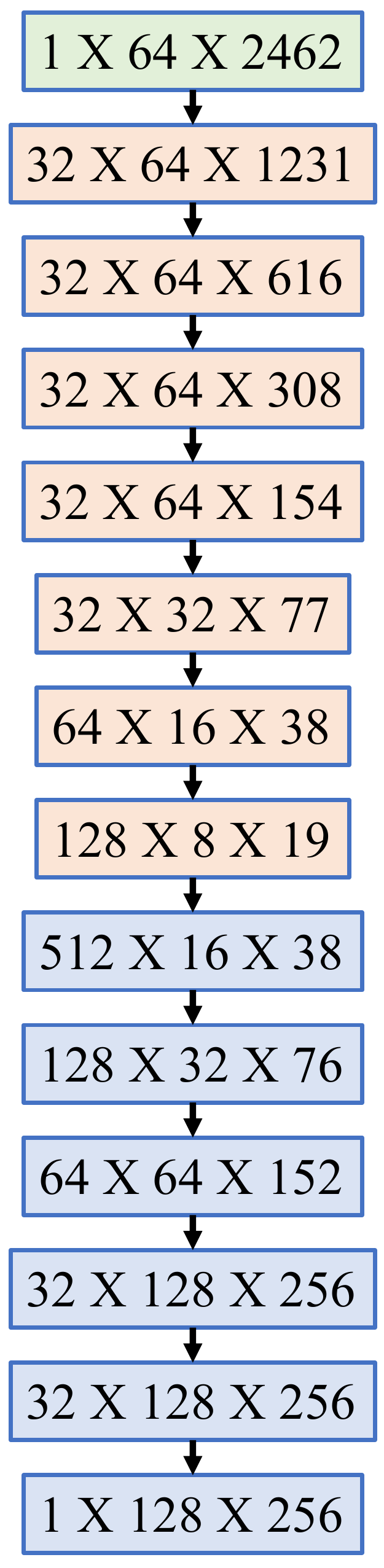}
    
    \caption{\label{fig:single_input_network}Base network setup, for handling a single transmitted plane wave signal. The green layer denotes the input layer. Orange layers are the encoding/downsampling layers. Blue layers are the decoding/upsampling layers. Most steps involve a decrease or increase of resolution by a factor of two, except for the last upsampling step which is a linear interpolation stage to adapt the aspect ratio.}
\end{figure}

Our base network architecture is depicted in Figure~\ref{fig:single_input_network}, and possesses an encoder-decoder or ``hourglass'' structure. In examining Figure~\ref{fig:single_input_network}, note the $C \times H \times W$ convention is used to describe the size of the layer's output: that is, number of channels by height by width.  The structure is referred to as hourglass due to its shape: in the initial ``encoding'' layers (shown in orange in Figure~\ref{fig:single_input_network}), the resolution decreases, i.e. $H$ and $W$ both decrease; while the number of channels increases, from the initial 1, to 32, 64, 128, and finally 512.  Thus, the layers get smaller but longer.  In the second ``decoding'' half of the network (shown in blue in Figure~\ref{fig:single_input_network}), the process is reversed: the resolution increases, while the number of channels goes down, finally reaching a single channel at the output.  Thus, the layers get larger but shorter. Note that due to the input geometry and output aspect ratio, one linear interpolation step is required. This results from the resolution increase not being a factor of two, going from a vertical resolution of 152 to 256.

On the encoding/downsampling path, the first four stages consist of a strided $3 \times 3$ convolution followed by batch normalization and a Relu operation.  Note that the stride used is 2 in the width dimension, which has the effect of downsampling that dimension by a factor of 2.  (In fact, the downsampling factor is sometimes not exactly 2; this is due to the nature of the convolutional padding used in the particular layer.)  The following three stages consist of an ordinary (non-strided) $3 \times 3$ convolution followed by batch normalization, Relu and a $2 \times 2$ maxpool.  The latter operation has the effect of reducing the resolution by a factor of 2 in both height and width dimensions (again, approximately, depending on the convolutional padding used).  For the decoding/upsampling path, each of the first three stages consists of a $3 \times 3$ convolution followed by batch normalization, Relu and a $\times2$ up-sampling.  The fourth stage involves a $3 \times 3$ convolution followed by batch normalization, Relu, and linear interpolation.  The fifth stage is a $3 \times 3$ convolution followed by batch normalization, Relu (and no upsampling/interpolation).  The final stage is a $1\times1$ convolution, which reduces the number of channels to one, and generates the output.

For the loss function, we used an $L_2$ loss comparing the output of the network to the expected sound speed map, based on down sampling and cropping the map the was used to generate the signal.

\begin{figure}
    \noindent
    \center
    \subfloat[Start]{
        \includegraphics[scale=0.21]{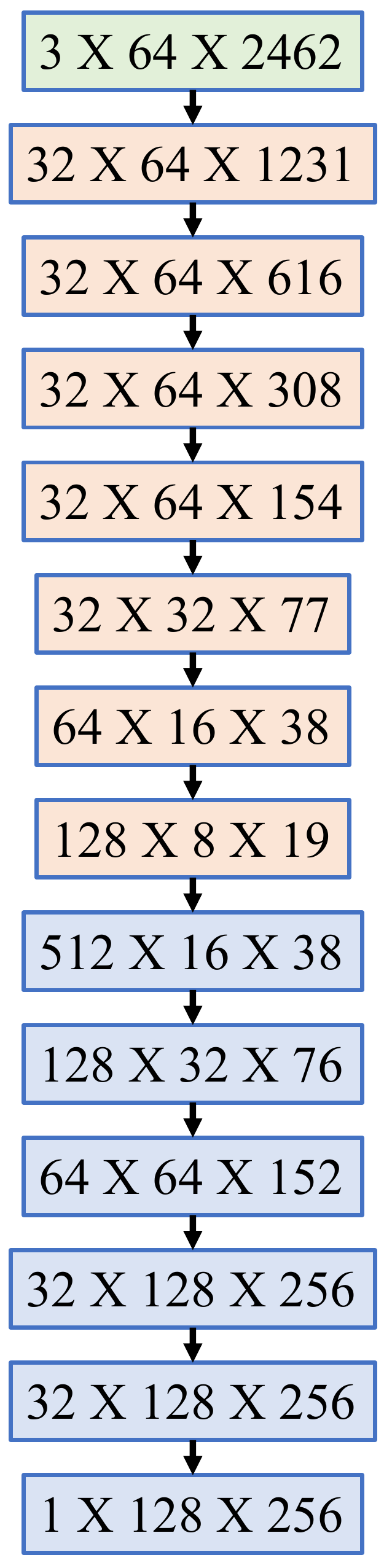}
    }
    \hfill{}%
    \subfloat[Middle]{
        \includegraphics[scale=0.21]{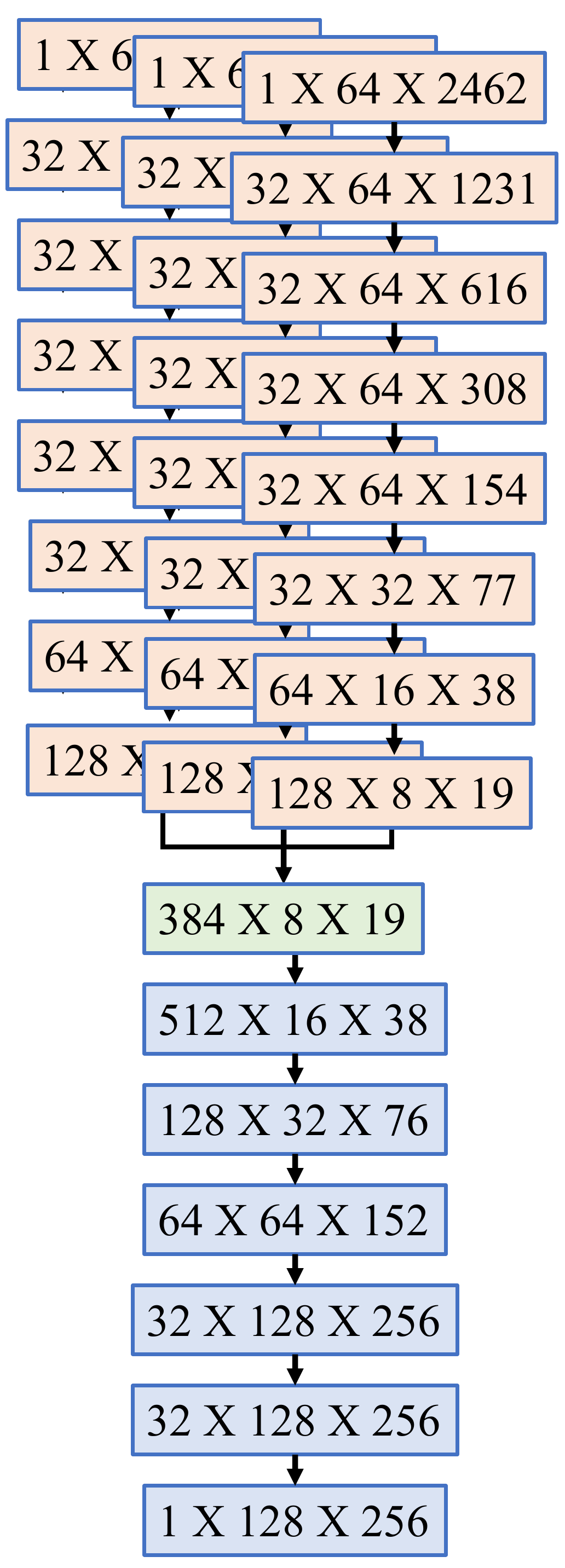}
    }
    \hfill{}%
    \subfloat[End]{
        \includegraphics[scale=0.21]{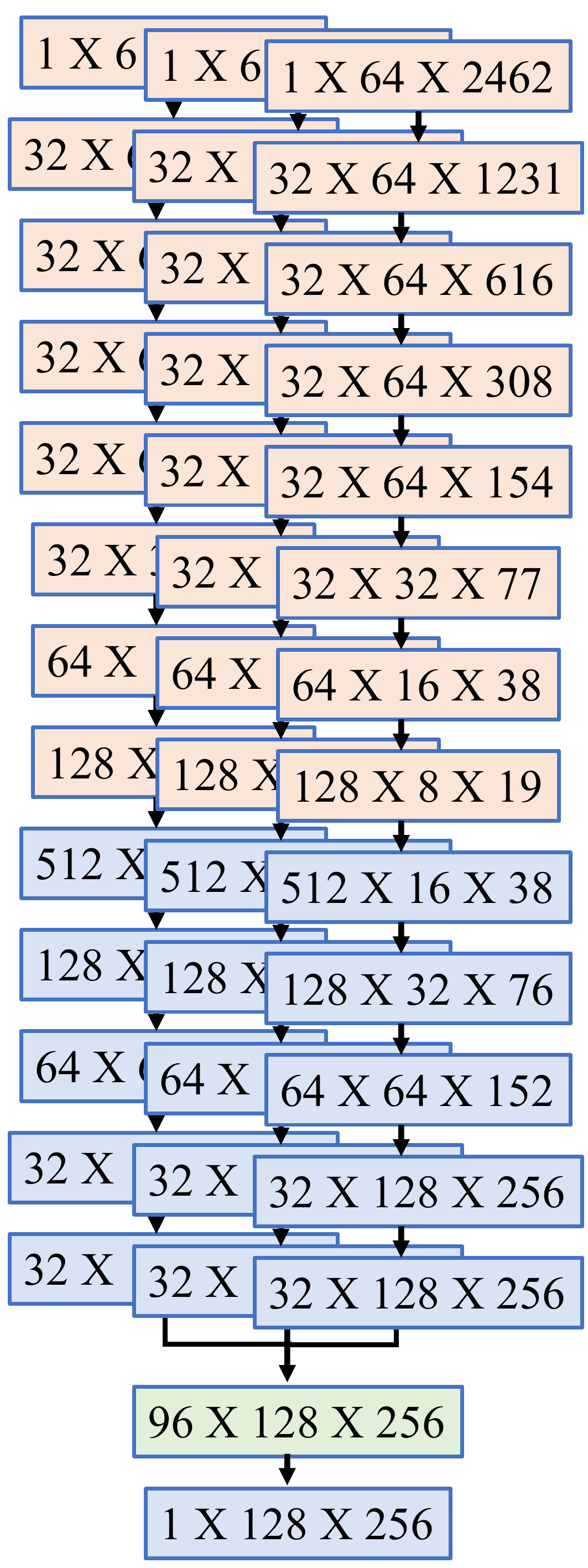}
    }
    
    \caption{\label{fig:mult_input_network}Network configurations for dealing with the data from multiple plane waves (multiple transmissions). The green layer denotes the data concatenation layer. Orange layers are the encoding/downsampling layers. Blue layers are the decoding/upsampling layers.}
\end{figure}

The base network, shown in Figure~\ref{fig:single_input_network}, has the capability of dealing with a single plane-wave.  We would like to use a variant of this base network for dealing with three plane-waves.  To that end, we test three different possibilities, as depicted in Figure~\ref{fig:mult_input_network}.  In the ``Start Network'', the three plane waves are simply concatenated into a 3-channel image, and the remainder of the base network is identical.  In the ``Middle Network'', the three plane-waves are each passed into identical subnetworks for the encoding/downsampling part of the base network; the results are then concatenated channel-wise, and the remainder of the decoding/upsampling part of the base network is the same.  In the ``End Network'', the same idea is used, but the channel-wise concatenation only happens at the very end, before the $1\times1$ convolution.  Note that for both Middle and End Networks, weight-sharing in the training phase ensures that each plane-wave is treated identically.

\section{Experimental Results}
\label{sec:results}

We present results for both synthetic data as well as initial results for real data.

For the training of the neural network, we used the k-wave toolbox for MATLAB to generate 6026 random training samples and 800 test samples using the procedure described in Sec~\ref{sec:simulations}. This took roughly two weeks using two NVIDIA GTX 1080i GPUs. Before feeding the data into the network, gain correction is applied at a rate of $0.48 \, dB / \mu s$ ($2.5 \, dB / cm$ at $1540\,m/s$). The channel data signals are then cropped in time to remove the transmit pulse, as depicted in Figure~\ref{fig:goal_a}. This is done to remove the transmit pulse that is several orders of magnitude stronger than the back-scattered signal, skewing signal statistics and results. Our physical system also suffers from electrical cross-talk in this temporal range during transmit, corrupting the data farther. All results presented were generated using the same network trained on the full simulated dataset. For training, random Gaussian noise and quantization noise were injected into the signal, which proved essential to avoid over training. Training was executed for 200 Epocs, based on the convergence of the loss on the test set.

\subsection{Results: Synthetic Data}

\begin{figure}
        \noindent
        \center
        \subfloat[Reference velocities]{
                \includegraphics[width=0.35\columnwidth]
                        {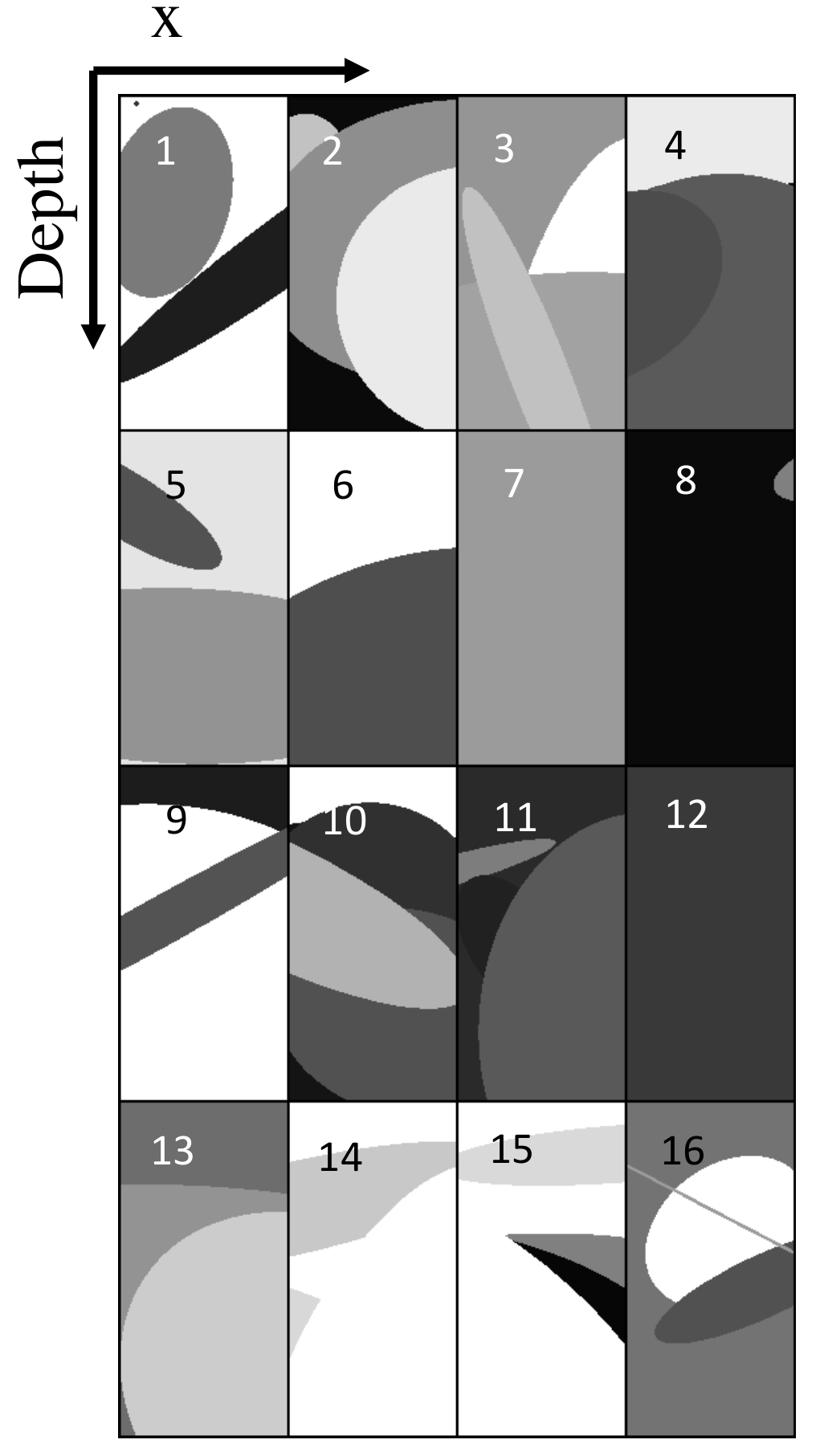}
        }%
        \subfloat[Middle plane wave]{
                \includegraphics[width=0.35\columnwidth]
                        {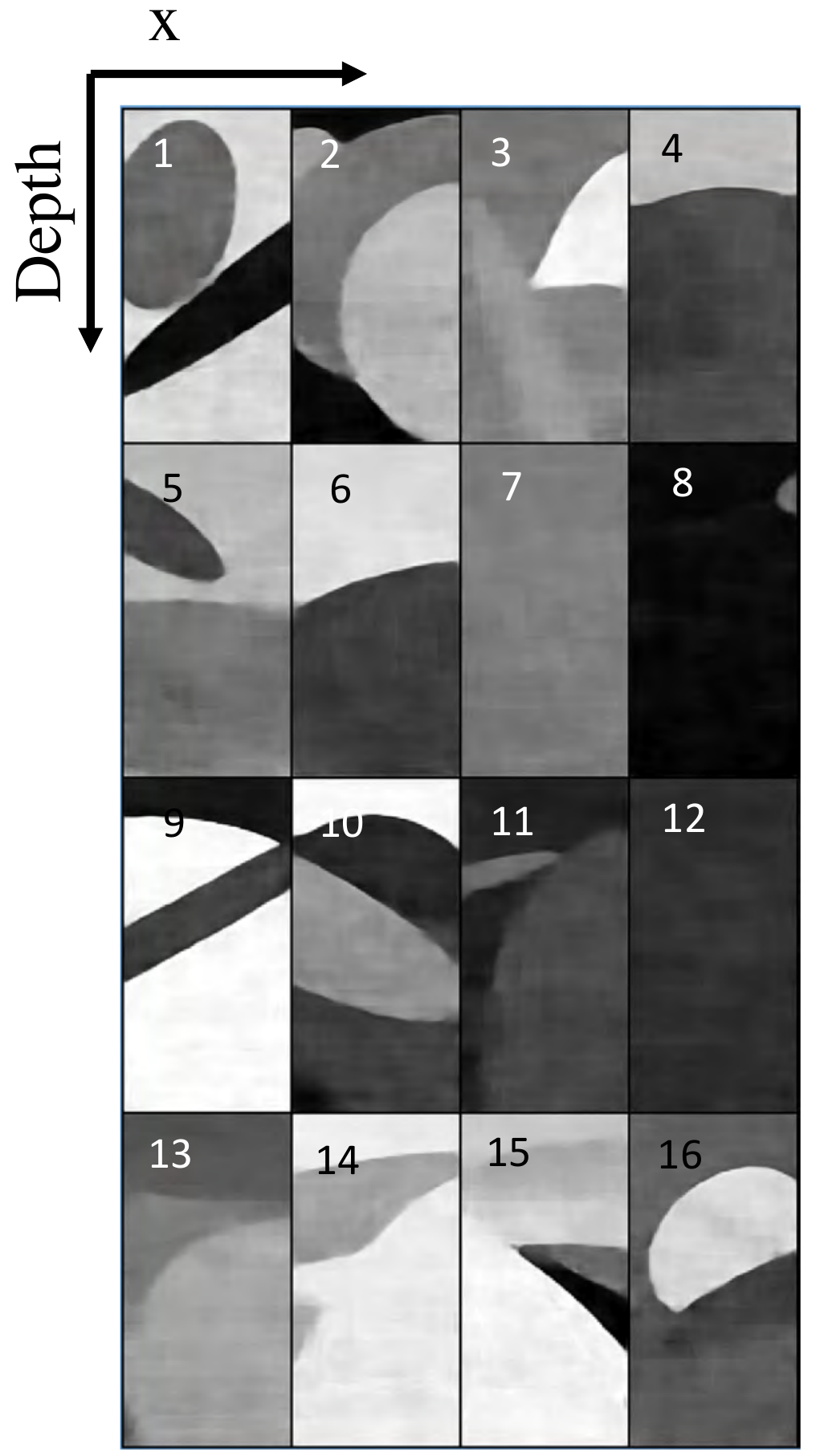}
        }

        \caption{\label{fig:sound_speed_recovery}Sound speed recovery maps on 16 test samples. Image (a) shows the ground truth data. Image (b) shows the sound speed maps recovered by the trained network using three plane waves and the ``middle'' network (see Figure~\ref{fig:mult_input_network}). Gray scale values are in the range of $1300\,m/s$ (black) to $1800\,m/s$ (white).}
\end{figure}

\begin{figure}
        \noindent
        \center
        \subfloat[Single plane wave]{
                \includegraphics[width=0.35\columnwidth]
                        {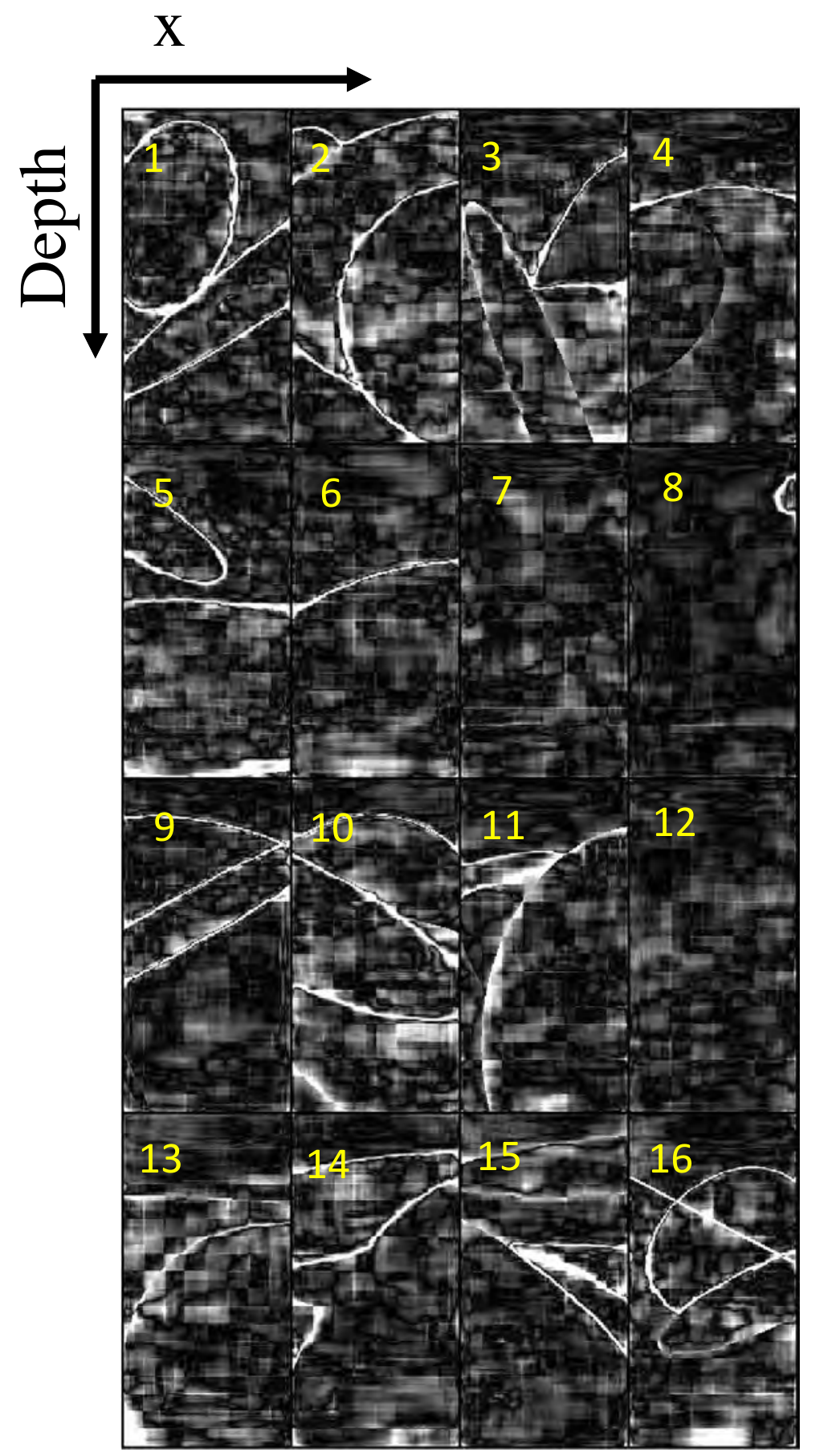}
        }%
        \subfloat[Three plane wave]{
                \includegraphics[width=0.35\columnwidth]
                        {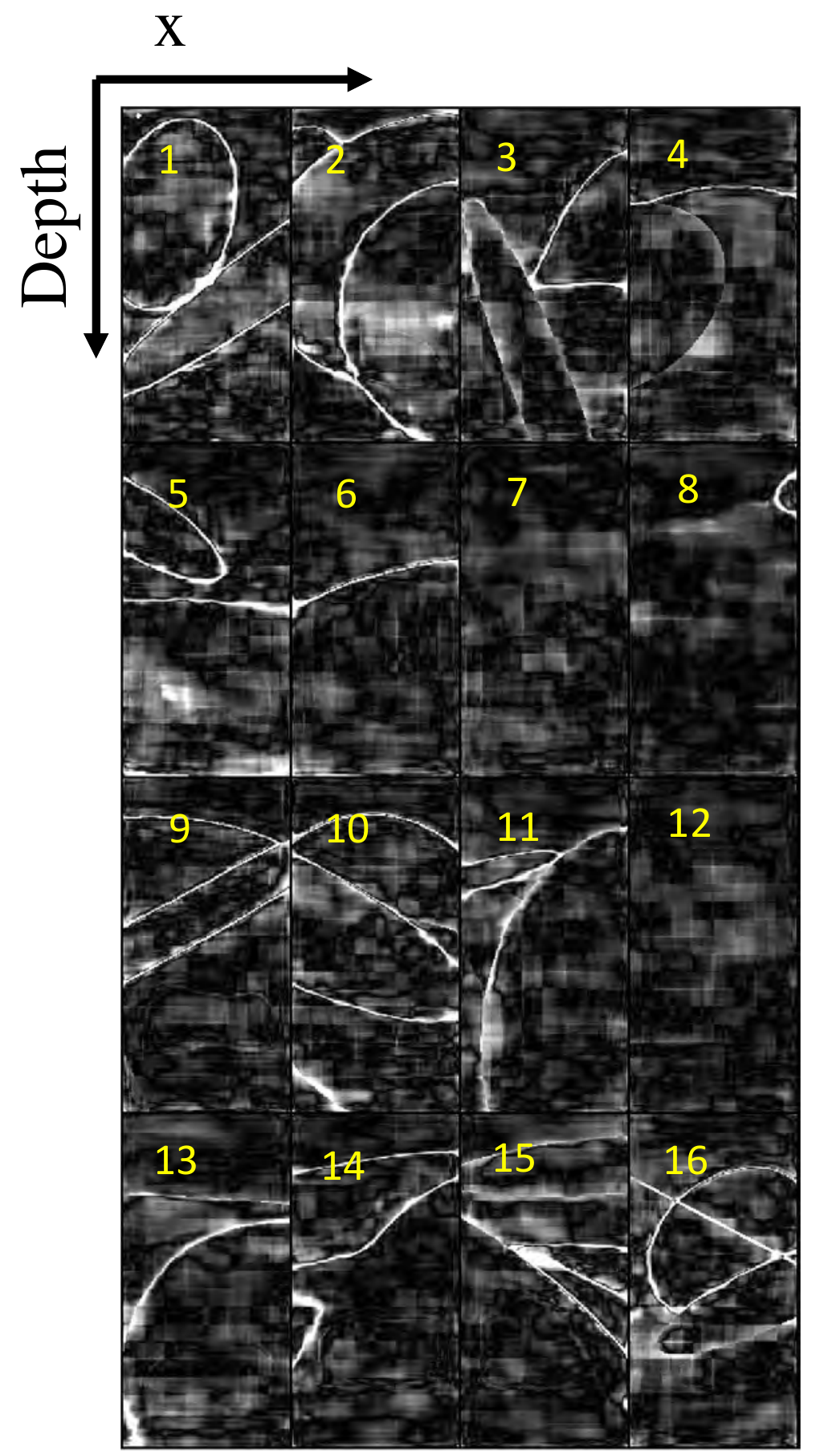}
        }

        \caption{\label{fig:abs_error}Absolute error on 16 test samples. Error has been cropped at $50\,m/s$ (white). Image (a) shows the results for the reconstruction using the single central plane wave. Image (b) shows the reconstruction using three plane waves and the ``Middle'' network  (see Figure~\ref{fig:mult_input_network})} 
\end{figure}

Fig.~\ref{fig:sound_speed_recovery} presents reconstruction results on several samples from the test data. Recovery works well on larger objects but can miss fine details, as can be seen for example in image 16. Fig.~\ref{fig:abs_error} shows absolute error values for the samples shown in Fig.~\ref{fig:sound_speed_recovery}, with a threshold at $50\,m/s$. As can be seen in frames 7 and 12, the system manages to recover sound speed also in the case of an empty domain, so information from speckle is also used and not just specular reflections. There is a slight misalignment at the edges, which is to be expected as even a tiny error in the location of the discontinuity or pixelization effects will cause a misalignment. Consequently, although we do show the classic root mean square error (RMSE) value, it does not convey the full story. The RMSE norm is a $L_2$ norm, making it sensitive to outliers. Thus we also report mean and median absolute errors, which are less sensitive to outliers.  Furthermore, to present error numbers that de-emphasize the issues due to localization around discontinuities, we further report the following modified error value: for each pixel, we take the minimum absolute error within a window with a radius of 5 pixels, for both the mean and median cases. For the mean error, in both cases, both the mean absolute error ($\mu$) as well as the standard deviation ($\sigma$) are reported. Results for all error measures are presented in Table~\ref{tab:error}. 

\begin{table*}
\caption{\label{tab:error}Reconstruction error for the train and test sets for our six recovery cases: single plane wave reconstruction for the three plane waves, and three plane wave reconstruction for the three joint reconstruction networks (Start, Middle, End). All values are in meters per seconds. RMSE measures the root mean square error. $\mu$ and $\sigma$ denote the mean and standard deviation values for the mean absolute error. Median shows the median absolute error. The star values are for our modified error value, taking the minimum absolute error over a window with a radius of 5 pixels, for both the mean and median error measures.}
\centering
\noindent
\begin{tabular}{|c|c|c|c|c|c|c|c|c|c|c|c|c|c|c|}
\hline 
Network  & \multicolumn{7}{c|}{Train } & \multicolumn{7}{c|}{Test}\tabularnewline
\hline 
 & RMSE  & $\mu$  & $\sigma$  & Median  & $\mu^{*}$  & $\sigma^{*}$  & Median$^{*}$ & RMSE  & $\mu$  & $\sigma$  & Median & $\mu^{*}$  & $\sigma^{*}$  & Median$^{*}$\tabularnewline
\hline 
\hline 
Left  & 22.4  & 14.6  & 17.0  & 10.6  & 2.0 & 4.6 & 0.18 & 24.8  & \cellcolor{lr}16.3  & 18.7  & \cellcolor{lr}11.8  & 2.6 & 6.0 & 0.22\tabularnewline
\hline 
Center  & \cellcolor{lr}23.3  & \cellcolor{lr}15.1  & \cellcolor{lr}17.8  & \cellcolor{lr}10.7  & \cellcolor{lr}2.5 & \cellcolor{lr}5.7 & \cellcolor{lr}0.19 & \cellcolor{lr}25.2  & 16.2  & \cellcolor{lr}19.3  & 11.4  & \cellcolor{lr}3.1 & \cellcolor{lr}7.0 & \cellcolor{lr}0.24\tabularnewline
\hline 
Right  & 19.2  & 12.2  & \cellcolor{lg}14.8  & 8.9  & 1.9 & 4.1 & 0.16 & 22.2  & 14.4  & 16.9  & 10.5  & 2.3 & 5.4 & 0.19\tabularnewline
\hline 
\hline 
Start  & 21.8  & 14.2  & 16.9  & 10  & 2.4 & 5.4 & \cellcolor{lr}0.19 & 24.3  & 15.6  & 18.6  & 11.0  & 2.9 & 6.5 & 0.23\tabularnewline
\hline 
Middle  & \cellcolor{lg}18.8  & \cellcolor{lg}11.5  & 14.9  & \cellcolor{lg}8.2  & 2.1 & 4.3 & 0.17 & \cellcolor{lg}20.5  & \cellcolor{lg}12.5  & \cellcolor{lg}16.1  & \cellcolor{lg}8.7  & 2.6 & 5.2 & 0.21\tabularnewline
\hline 
End  & 18.9  & 11.9  & \cellcolor{lg}14.8  & 8.5  & \cellcolor{lg}1.6 & \cellcolor{lg}3.7 & \cellcolor{lg}0.14 & 20.8  & 12.9  & 16.3  & 9.0  & \cellcolor{lg}2.0 & \cellcolor{lg}5.0 & \cellcolor{lg}0.16\tabularnewline
\hline 
\end{tabular}
\end{table*}

Available research suggests that for clinically relevant results, measurement accuracy on the order of $30\,m/s$ is useful. All results are well within that range, and error measures that account for outliers at edges are an order of magnitude better. While more work is required to improve results on real data, we see very strong potential with our proposed technology.

We now turn to the issue of multiple inputs, that is, the use of multiple plane waves in image formation.  While combining multiple inputs at the first layer (``Start Network'') does not improve reconstruction results, combining in both the middle (``Middle Network'') and the end (``End Network'') does provide some improvement, as can be seen, Table~\ref{tab:error} and Figure~\ref{fig:abs_error}. However all cases are close to the recovery limit, so we expect more value in terms of stability to noise when dealing with real data.

\subsection{Results: Real Data}

\begin{figure}
    \noindent
    \center
    \subfloat[B-mode image]{
    \label{fig:poly-a}
        \includegraphics[width=0.36\columnwidth]{poly_b_mode}
    }~
    \subfloat[Ground truth sound speed]{
        \label{fig:poly-b}
        \includegraphics[width=0.43\columnwidth]{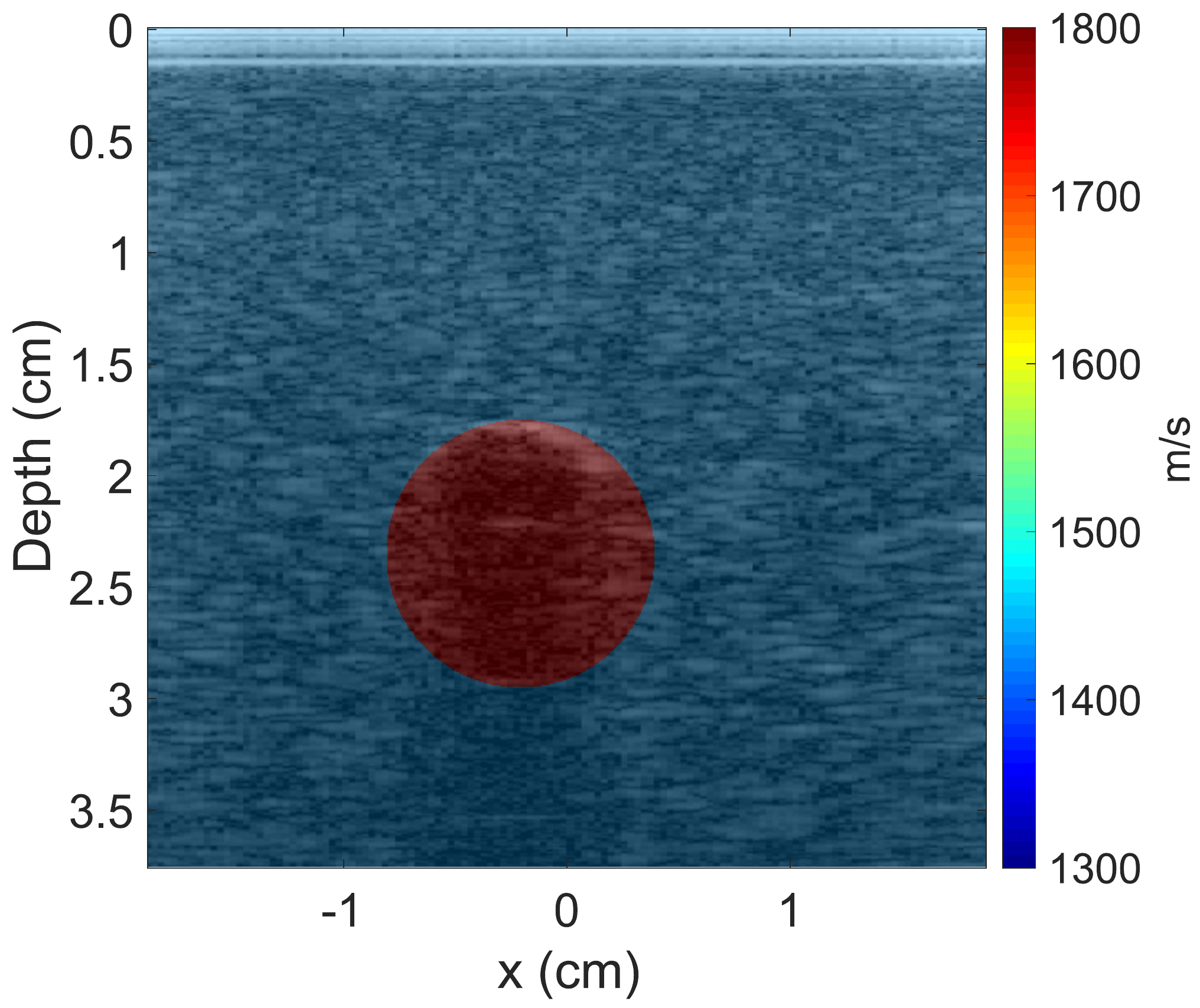}
    }
    
    \subfloat[sound speed - single plane wave]{
        \label{fig:poly-c}
        \includegraphics[width=0.44\columnwidth]{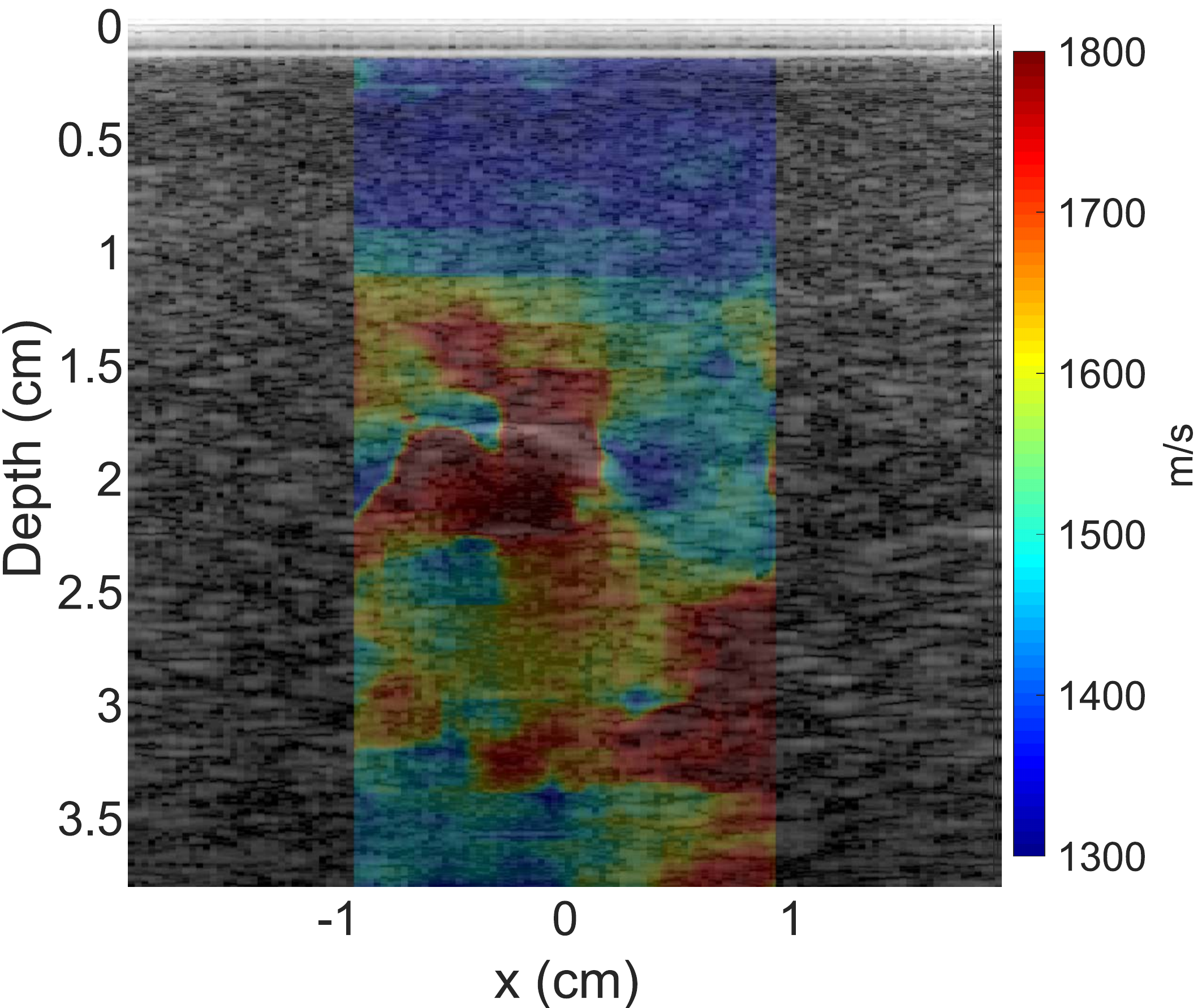}
    }
    \subfloat[sound speed - three plane waves]{
        \label{fig:poly-d}
        \includegraphics[width=0.44\columnwidth]{poly_speed}
    }
    
    \caption{\label{fig:poly}Polyurethane phantom with  inclusion. (a) shows the B-mode image. (b) shows ground truth sound speed map, measured at $1440\,m/s$ for background and $1750\,m/s$ for the inclusion. (c) shows the sound speed recovery from a single  plane wave. (d) shows the sound speed recovery for three plane waves with the ``Middle'' network}
\end{figure}

\begin{figure}
    \noindent
    \center
    \subfloat[B-mode image]{
           \label {fig:thyroid-a}
        \includegraphics[width=0.39\columnwidth]{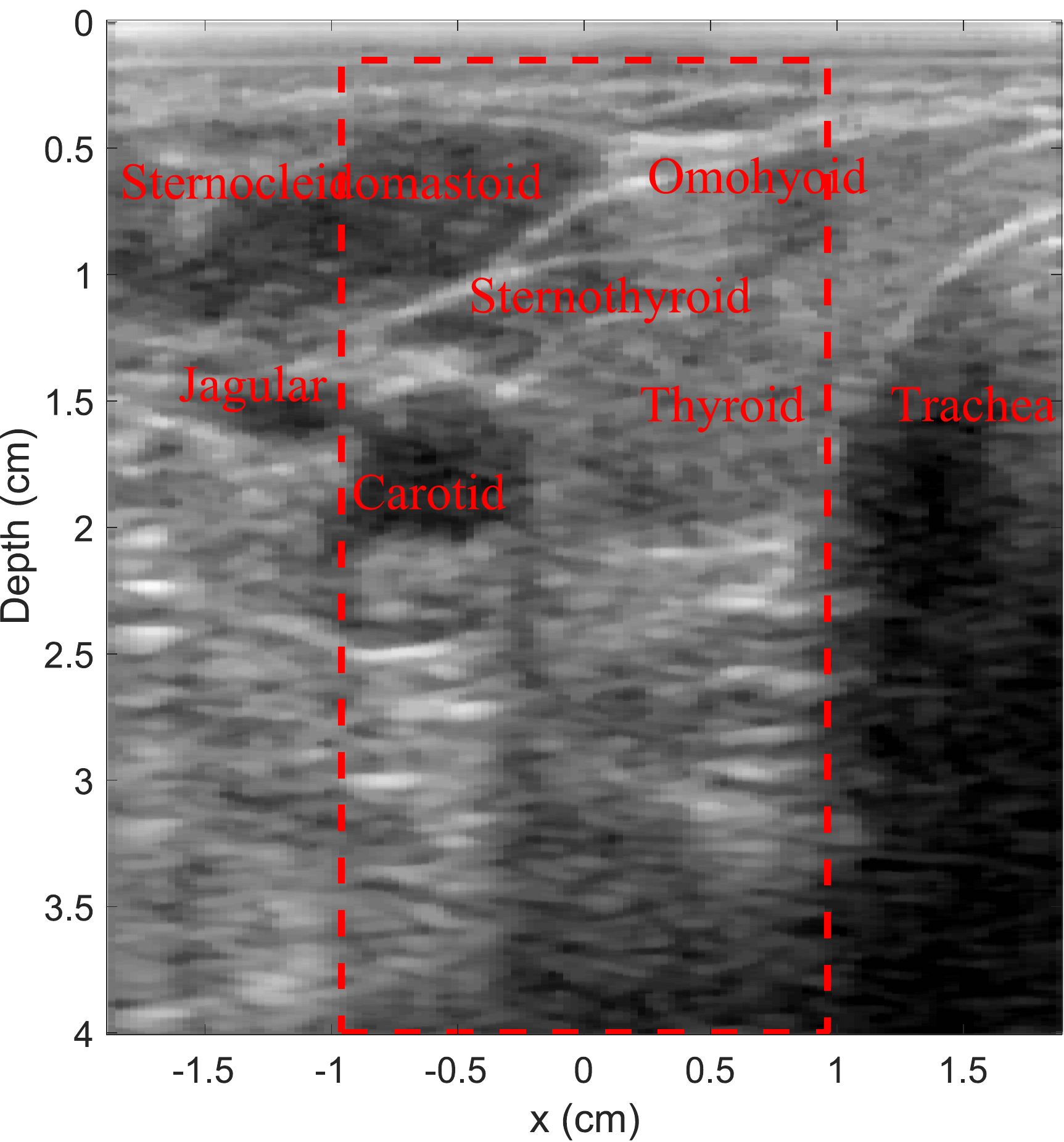}
    }
    \subfloat[Neck anatomy]{
           \label {fig:thyroid-b}
        \includegraphics[width=0.37\columnwidth]{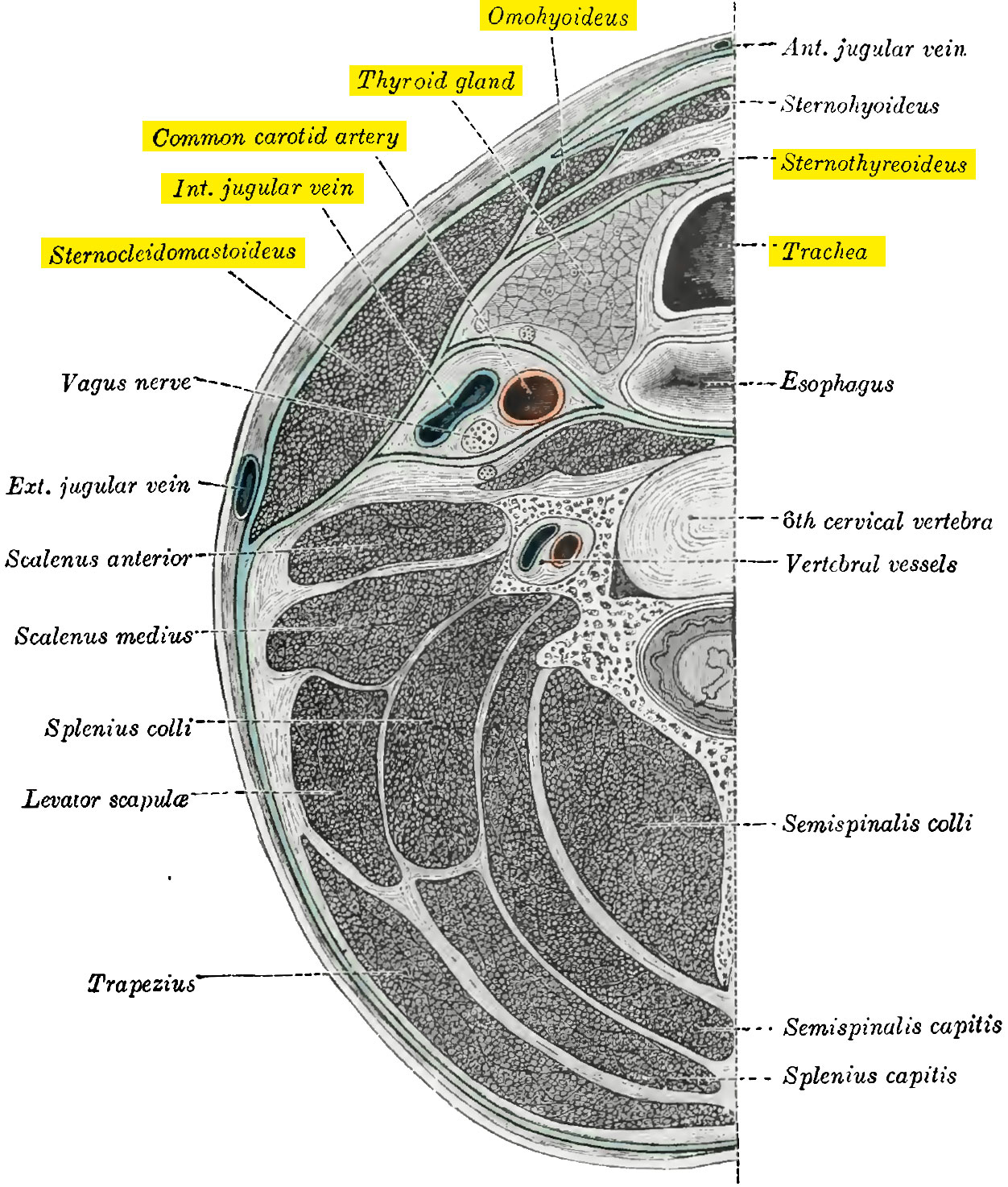}
    }
    
    \subfloat[Sound speed - single plane wave]{
           \label {fig:thyroid-c}
        \includegraphics[width=0.44\columnwidth]{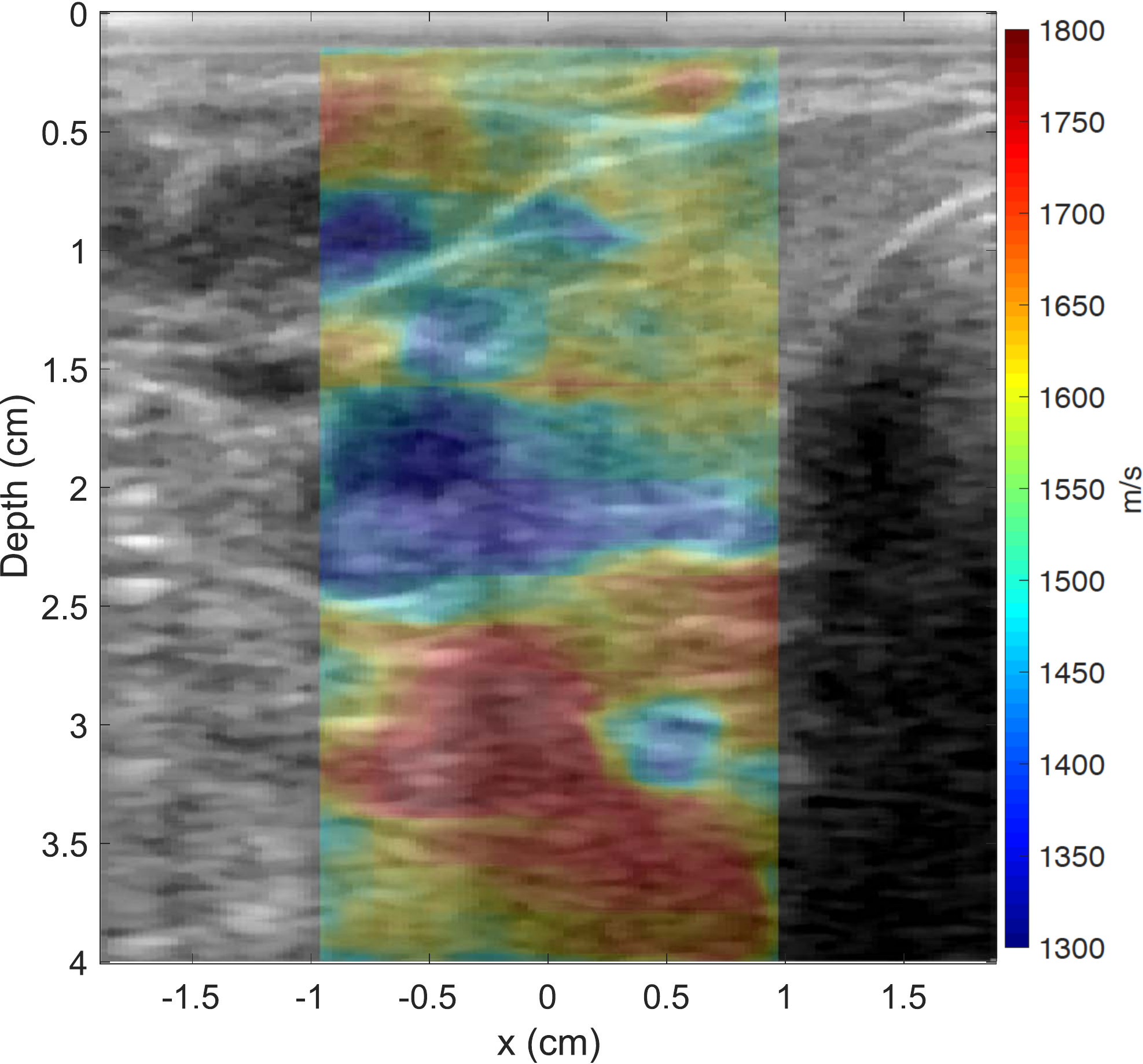}
    }
    \subfloat[Sound speed - three plane waves]{
           \label {fig:thyroid-d}
        \includegraphics[width=0.44\columnwidth]{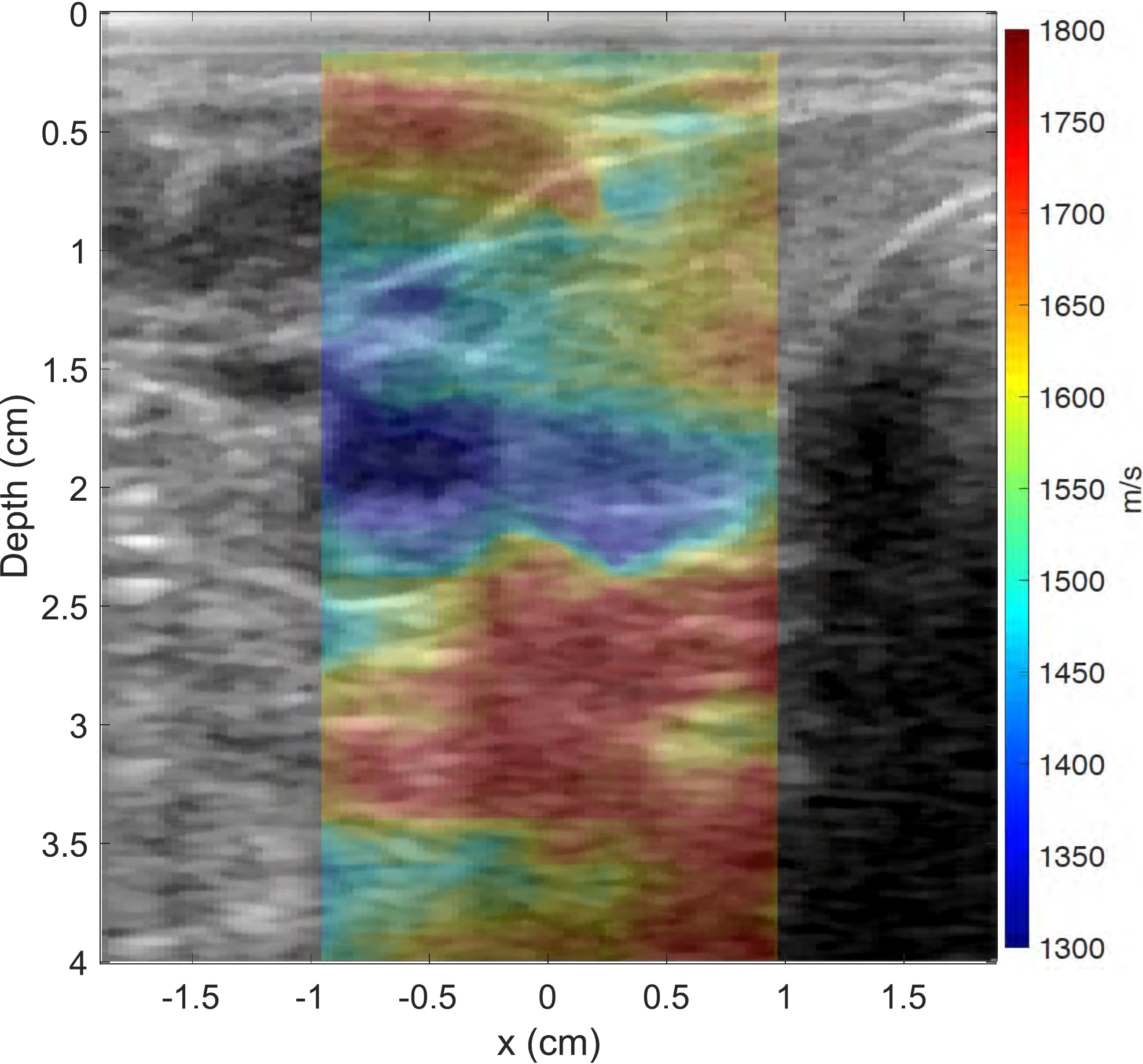}
    }
    
    \caption{\label{fig:thyroid}Sound speed recovery for the neck. Image (a) shows a b-mode US image reconstruction. Image (b) shows an anatomical drawing of a cross section of the neck \cite{grayAnatomyHumanBody1918}, with the anatomical landmarks appearing in (a) highlighted. Image (c) shows the sound speed reconstruction using the single central plane wave. Image (d) shows the sound speed reconstruction using three plane waves and the ``Middle'' network}
\end{figure}

\begin{figure}
    \noindent
    \center
    \subfloat[B-mode image]{
        \label{fig:leg-a}
        \includegraphics[width=0.35\columnwidth]{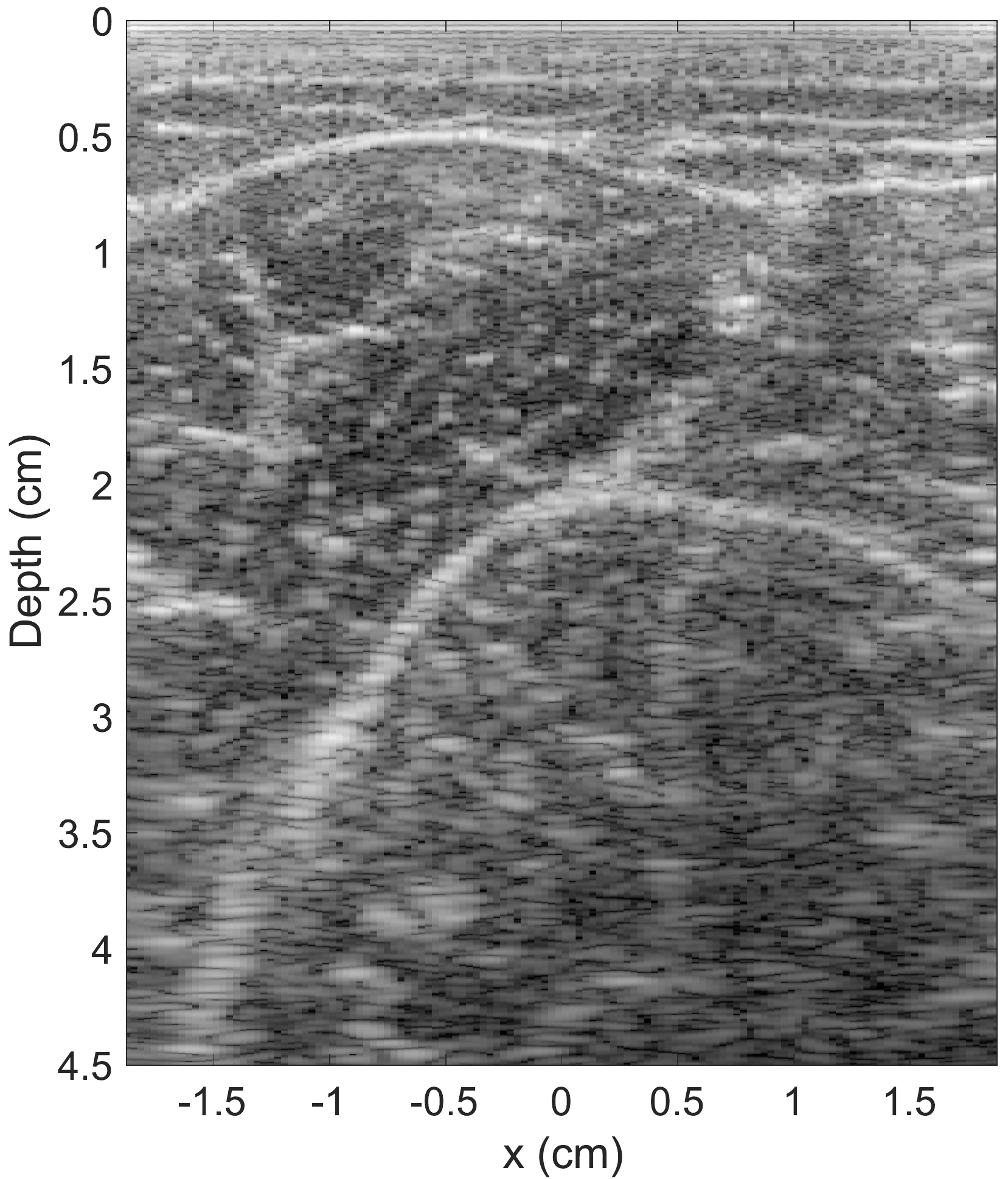}
    }
    \subfloat[Lower leg anatomy]{
            \label{fig:leg-b}
        \raisebox{3mm}{
            \includegraphics[width=0.50\columnwidth]{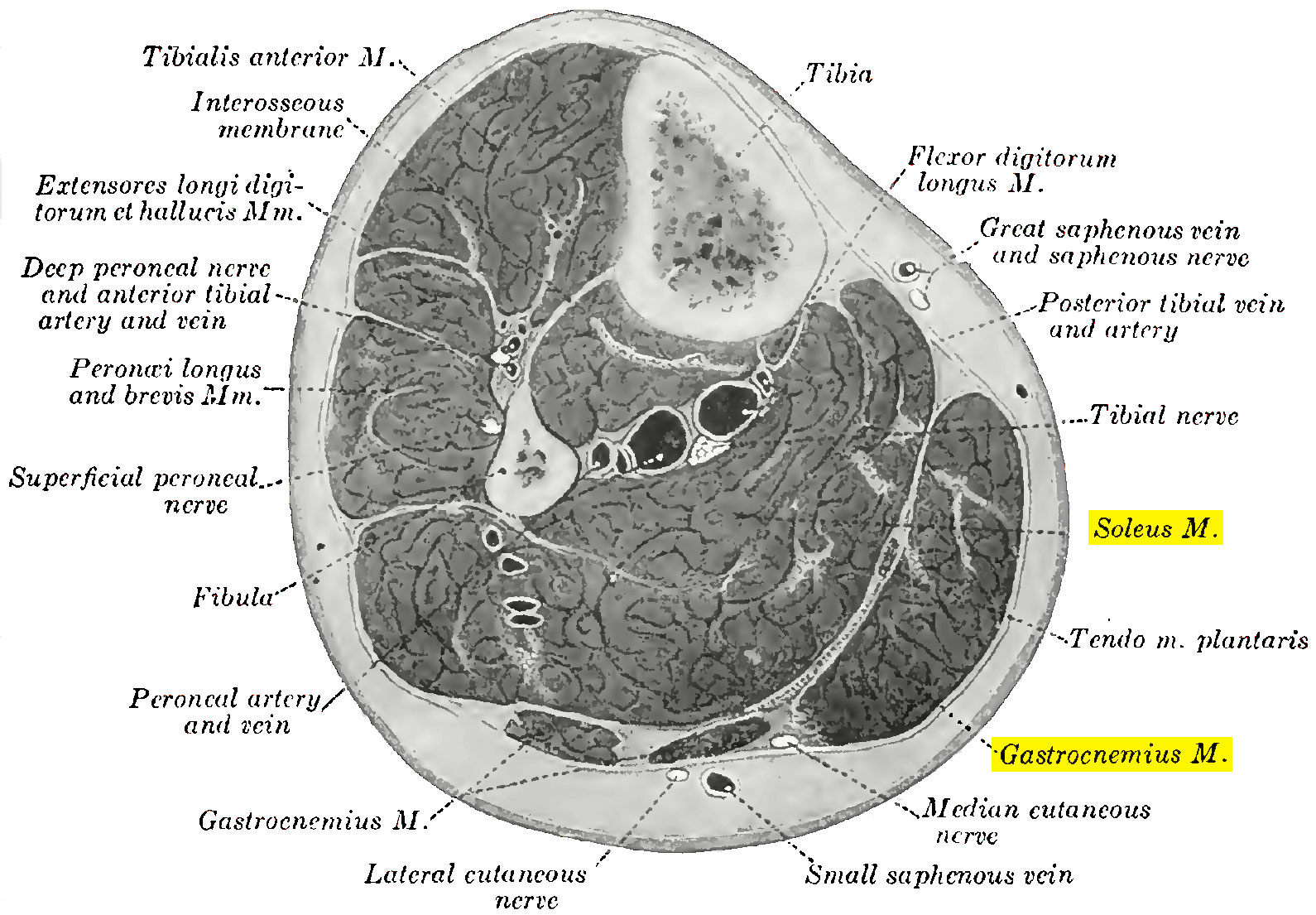}
            }
    }
    
    \subfloat[Sound speed, straight leg]{
            \label{fig:leg-c}
        \includegraphics[width=0.43\columnwidth]{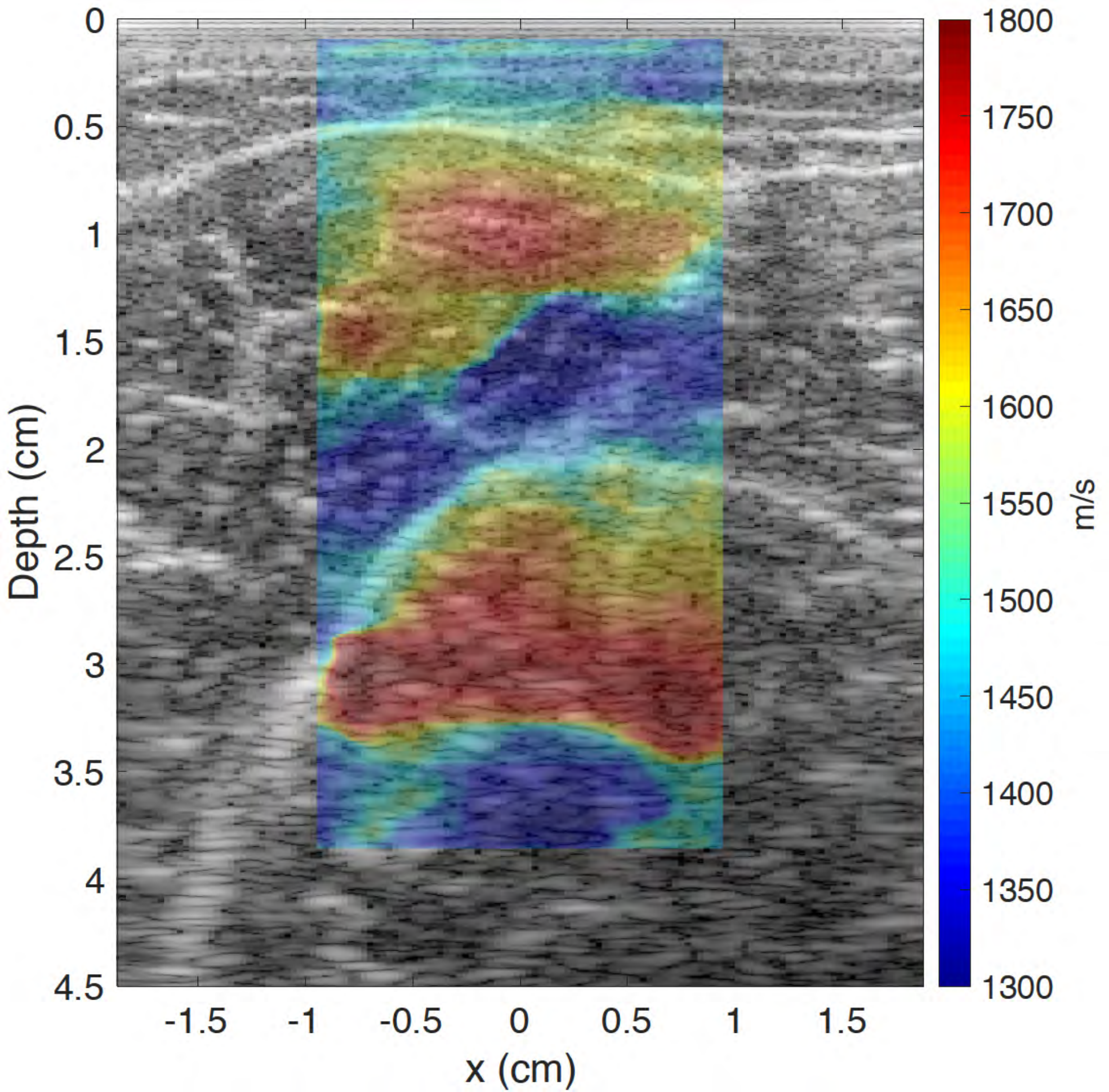}
    }
    \subfloat[Sound speed, Bent leg]{
            \label{fig:leg-d}
        \includegraphics[width=0.43\columnwidth]{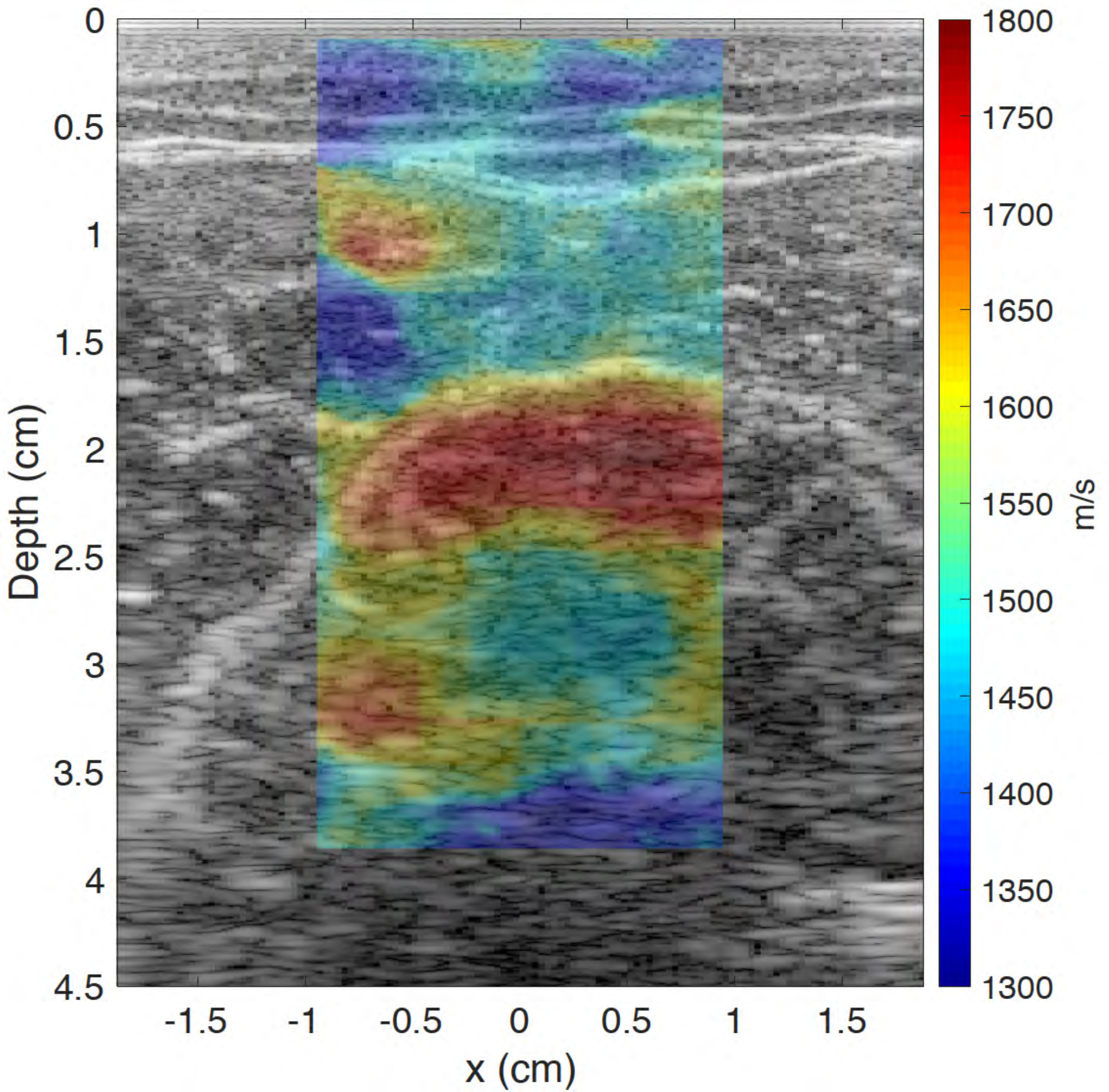}
    } 
    
    \caption{\label{fig:leg}Sound speed imaging of the lower leg muscle. Image (a) shows the b-mode US image reconstruction with the major muscles of interest delineated (gastrocnemius and soleus). Image (b) shows an anatomical sketch of a cross section of the lower leg \cite{grayAnatomyHumanBody1918}. Images (c) and (d) show sound speed reconstruction with toes in flexion, with a straight leg on the left, where we expect the gastrocnemius to be active, and a bent leg on the right, where we expect the soleus to be active.}
\end{figure}

For the case of real data, we look at three data-sets: (1) a polyurethane phantom with an inclusion (Fig.~\ref{fig:poly}), (2) a cross section of the neck  (Fig.~\ref{fig:thyroid}) and (3) an image of the calf muscles (gastrocnemius and soleus, Fig.~\ref{fig:leg}). All data was collected using a Cephasonics ultrasound system using a 128 linear probe transmitting at $5\,MHz$. Both human scans were taken as part of an MIT Committee on the Use of Humans as Experimental Subjects (COUHES) approved protocol.

\begin{figure}
    \noindent
    \center
    \subfloat[B-mode image]{
        \includegraphics[width=0.42\columnwidth]{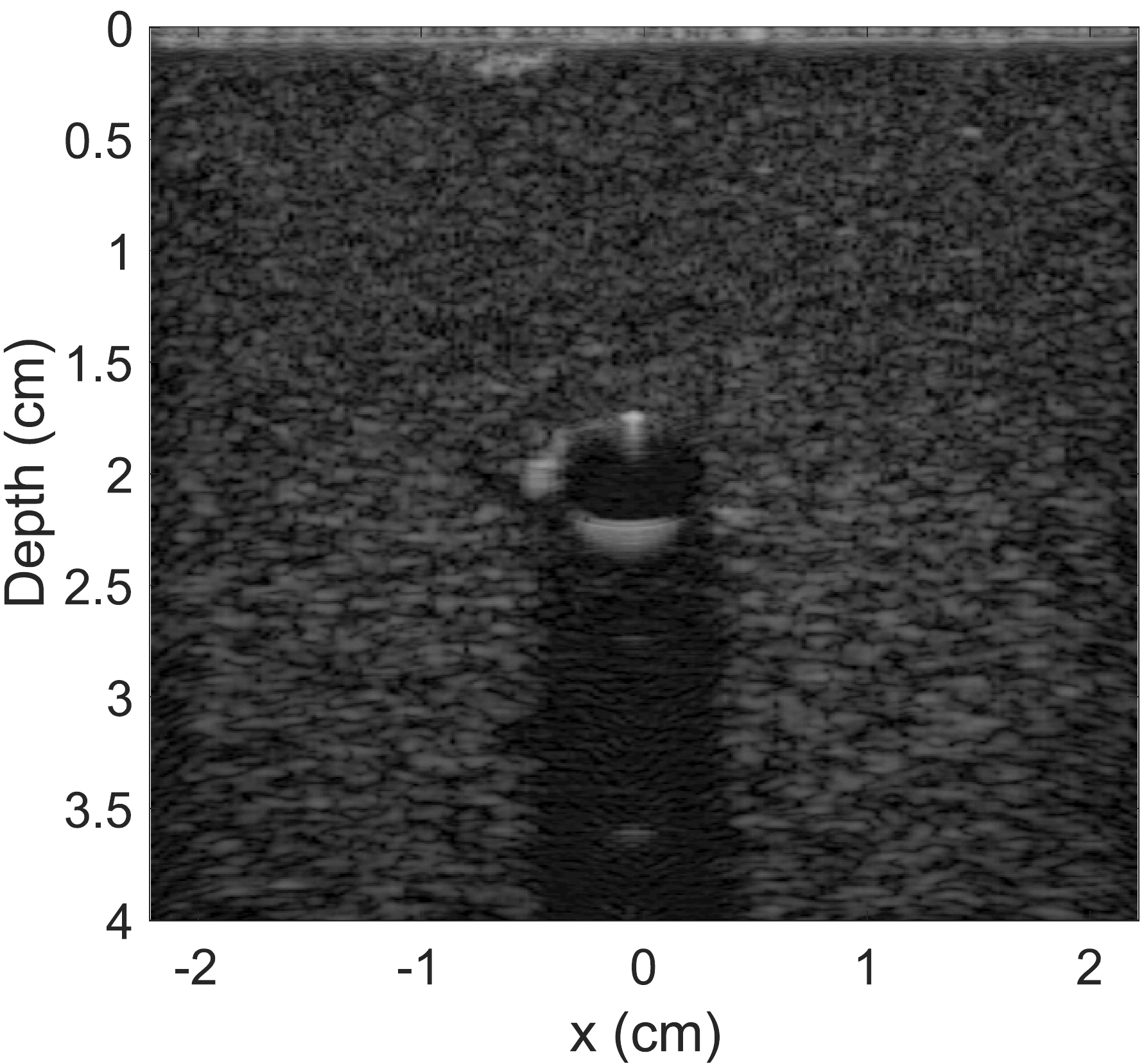}
    }
    \subfloat[Shear wave sound speed]{
        \includegraphics[width=0.47\columnwidth]{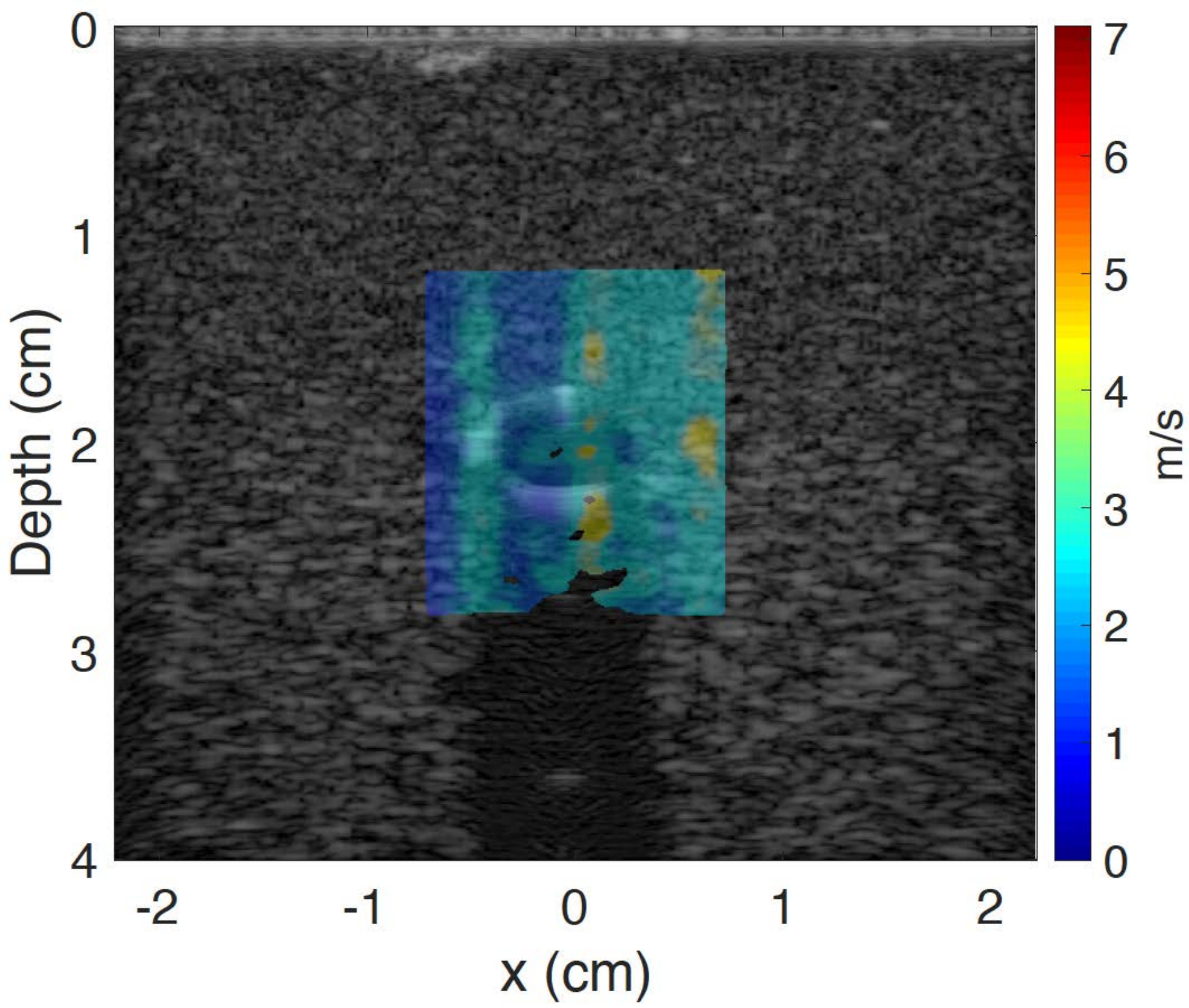}
    }

    \caption{\label{fig:poly-shear}Shear wave imaging of the polyurethane phantom presented in Fig.~\ref{fig:poly}. Image (a) shows the b-mode image and image (b) shows the overlaid shear wave sound speed map.}
\end{figure}

\begin{figure}
    \noindent
    \center
    \subfloat[B-mode image]{
        \includegraphics[width=0.41\columnwidth]{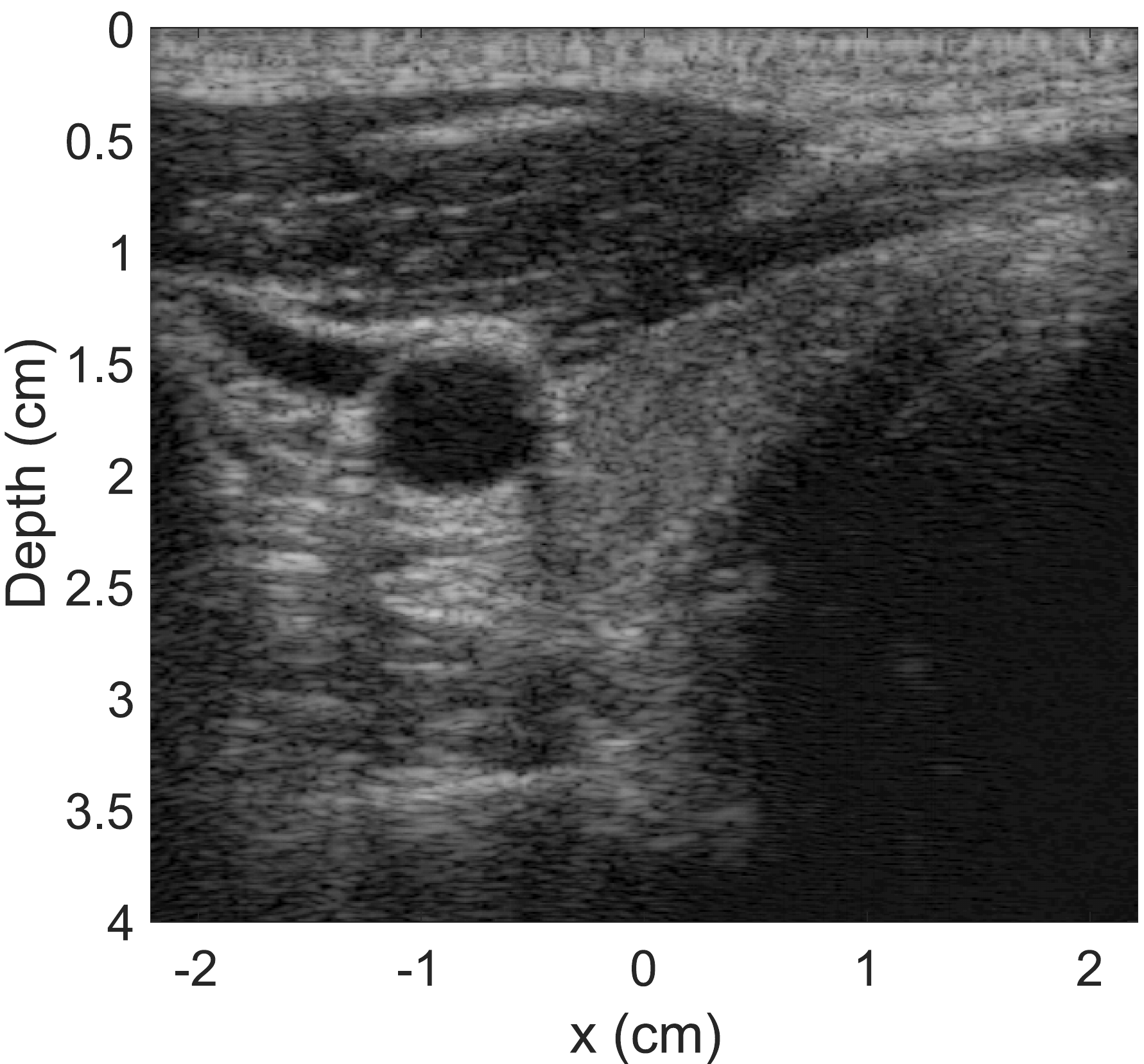}
    }
    \subfloat[Shear wave sound speed]{
        \includegraphics[width=0.45\columnwidth]{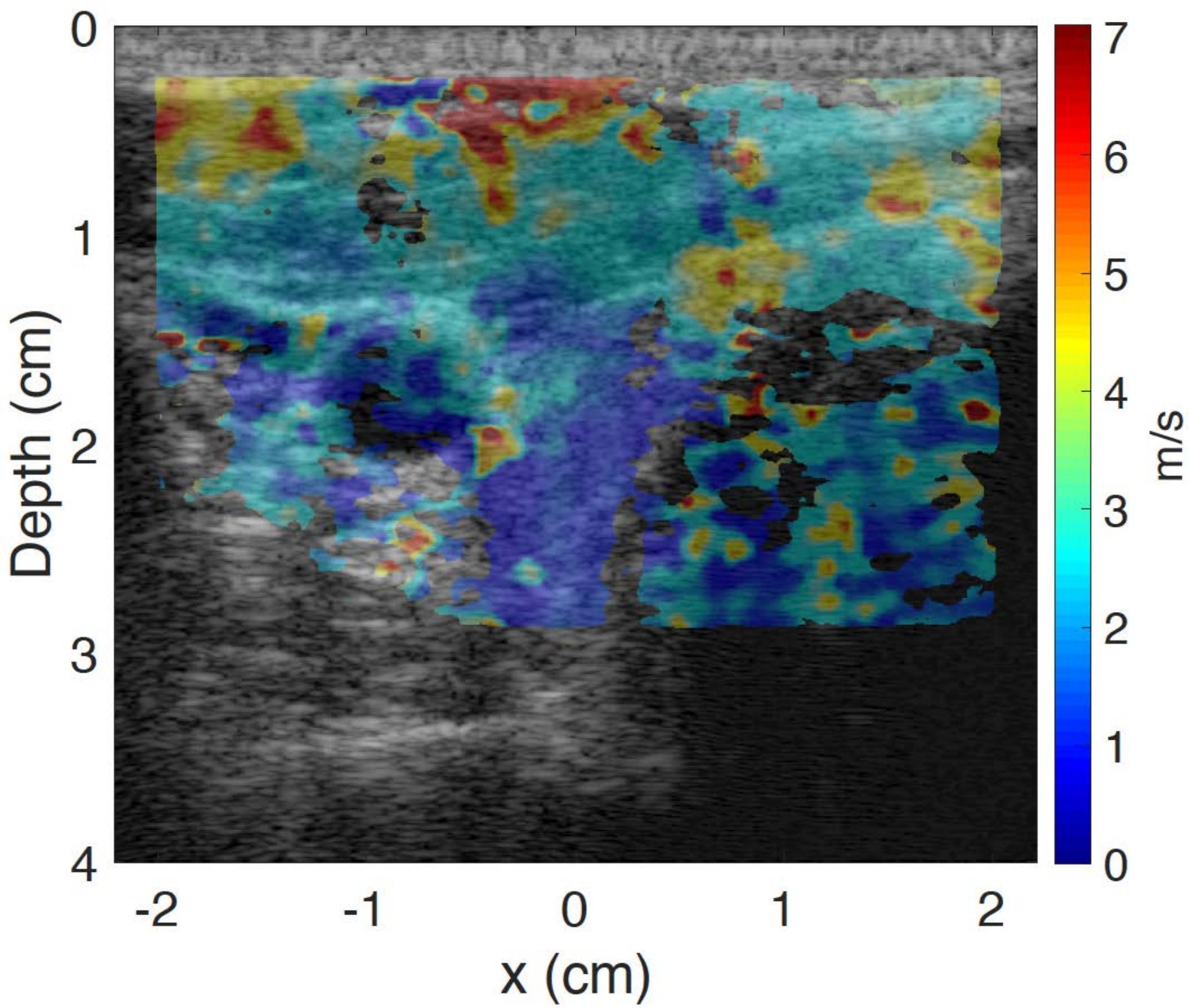}
    }

   \caption{\label{fig:thyroid-shear}Shear wave imaging of the neck, taken from the same angle of view presented in Fig.~\ref{fig:thyroid}. Image (a) shows the b-mode image, image (b) shows the overlaid shear sound speed map.}
\end{figure}

\begin{figure}
    \noindent
    \center
    \subfloat[Straight leg]{
        \includegraphics[width=0.43\columnwidth]{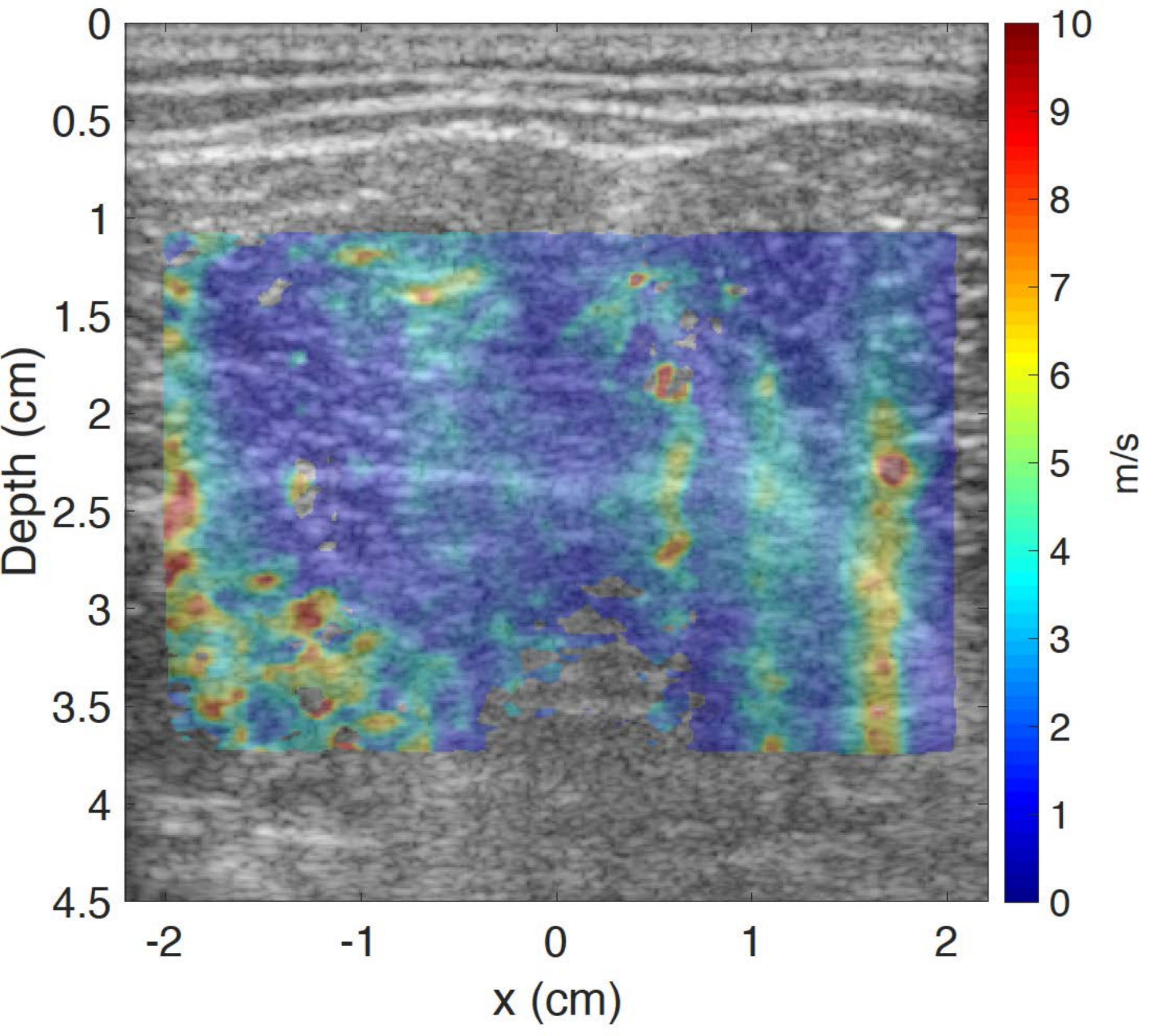}
    }
    \subfloat[Bent leg]{
        \includegraphics[width=0.43\columnwidth]{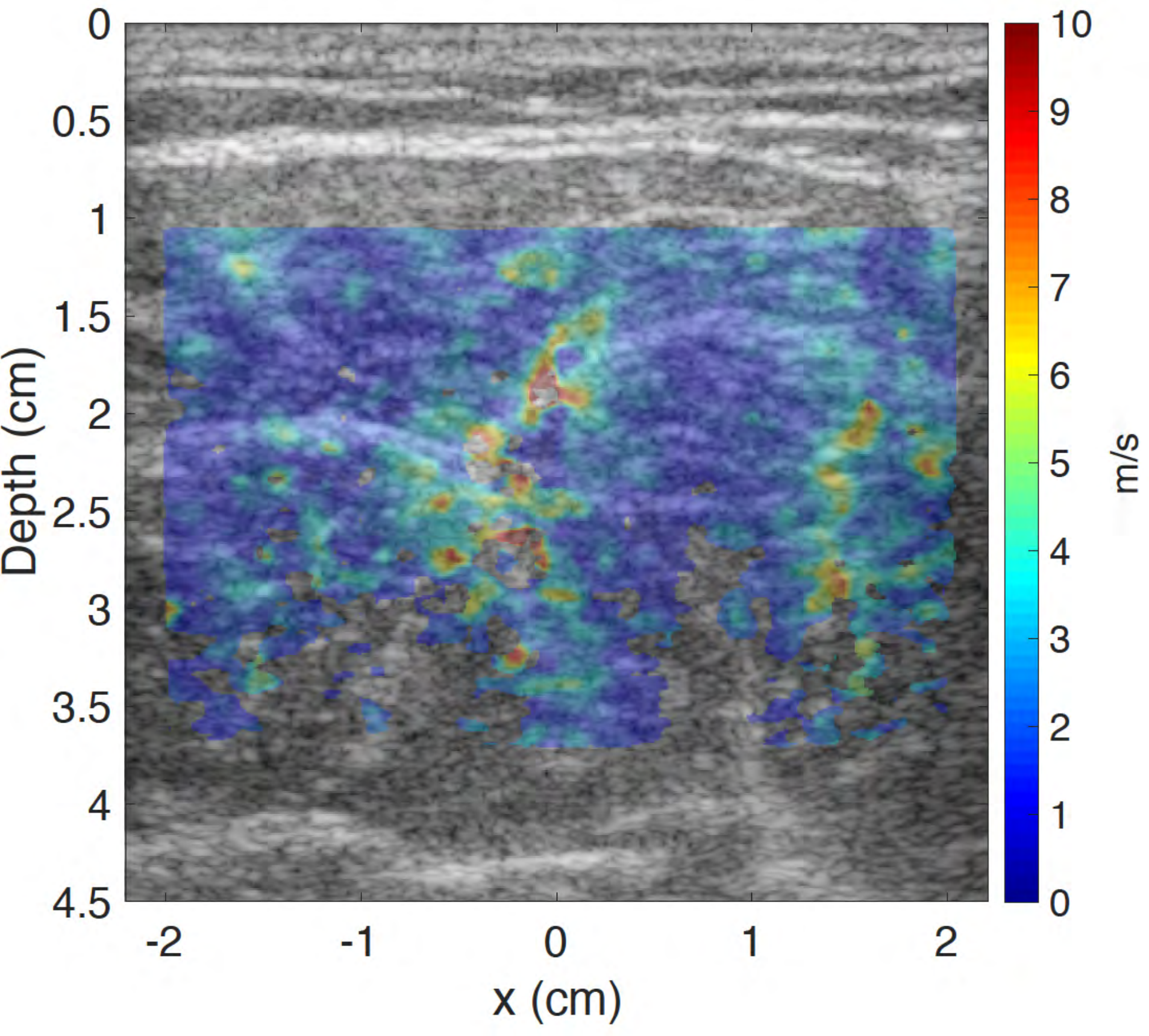}
    }
    
    \caption{\label{fig:leg-shear}Shear wave imaging of the leg matching Figs~\ref{fig:leg-c} and \ref{fig:leg-d}.}
\end{figure}

In all cases we show the results for the sound speed map reconstruction using a single plane wave, as well as three plane waves using the ``middle'' network. Results using the ``middle'' and ``end'' are very similar, so for the sake of brevity, we omit the output of the ``end'' network. In addition to the pressure wave sound speed images, we also collected shear wave sound speeds at the same location for comparison. Results for the polyurethane phantom are given in Fig.~\ref{fig:poly-shear}, the neck in Fig.~\ref{fig:thyroid-shear} and the calf in Fig~\ref{fig:leg-shear}. Shear wave data were collected using a GE Logic E9 system with a GE 9L 192 element linear probe.

For the polyurethane phantom we have a ground truth sound speed map measured based on transmission travel time and presented in Fig~\ref{fig:poly-b}. Background sound speed is $1440\,m/s$ and inclusion sound speed is $1750\,m/s$. Comparing to the sound speed field reconstruction using a single plane wave (Fig~\ref{fig:poly-c}), we see that the near (top) side of the inclusion is detected correctly (with a slight sound speed overshoot) but the bottom half is not. Sounds speed close to the probe is under-estimated. Closer to the inclusion, sound speed is over-estimated, with large artifacts deeper into the phantom. In contrast, the three plane wave reconstruction shows significantly better results. The inclusion is fully detected, with an accuracy that is barely possible on the b-mode image. There are significantly fewer artifacts as well, but sound speed is still underestimated closer to the probe and overestimated close to the inclusion, although not as much as for the single plane-wave case.

The shear wave sound speed map presented in Fig~\ref{fig:poly-shear} shows that shear wave imaging does not fair well with this phantom. Although we do not have ground truth shear wave sound speed maps, it is easy to see that the inclusion is not detected at all, and the sound speed map suffers from vertical artifacts, making the quality of these results questionable.

For the neck cross-section sample, the annotated b-mode image is presented in Fig.~\ref{fig:thyroid-a} with the matching anatomical sketch in Fig.~\ref{fig:thyroid-b}. We do not have a ground truth sound speed map, in this case, to compare to, but we do see that the recovered sound speed map follows the anatomy, as well as the shear wave image, differentiating between muscle, carotid artery and thyroid gland. The sound speed inside the carotid is underestimated, although we suspect that is due to the lack of backscatter energy from the blood content. Additionally, the recovered sound speed map for the near muscles (sternocleidomastoid, omohyoid, and sternothyroid), as well as the thyroid, match the expected statistical values. The deeper muscles are differentiated correctly anatomically but the sound speed does is overestimated. In this, case, probably due to the higher feature density, there is a much smaller difference between the single plane-wave and three plane-wave versions. Turning our attention to the shear wave sound speed image presented in Fig.~\ref{fig:thyroid-shear}, we see that there is a general (though by no means perfect) correlation between the sound speed field generated by our network and the shear wave speed field. This correlation represents a general validation of our technique.

A cross-section scan of the calf muscles, specifically the gastrocnemius and soleus, is shown in Fig.~\ref{fig:leg}. Fig.~\ref{fig:leg-a} shows the annotated b-bode image and Fig.~\ref{fig:leg-b} shows an anatomical sketch. In this example, we explore functional imaging. Both sound speed maps are taken with the toes in flexion and under a small load to activate the calf muscles. The first frame, presented in Fig.~\ref{fig:leg-c}, shows an image with a straight leg, where we expect the gastrocnemius (external muscle) to be the main active muscle. The second frame, shown in Fig.~\ref{fig:leg-d}, shows the results with a bent leg, where we expect the soleus (internal muscle) to be the one doing most of the work. The contraction of the gastrocnemius muscle is very obvious, although in the case where it is contracting, it appears that the lower half is estimated as having a low sound speed instead of high. The response of the soleus is not as obvious, as sound speed estimation is too high in both cases, but can still be observed in the results. Assessed sound speed in the relaxed gastrocnemius muscle in Fig.~\ref{fig:leg-d} is about $1540 m/s$ and of the subcutaneous fat around $1450 m/s$, both of which are extremely close to the expected values.

As before, shear wave sound speed images are presented for both cases in Fig~\ref{fig:leg-shear}. As is easily seen, these results do not provide any meaningful information. It is our general experience that full frame shear wave sound speed images on loaded muscles tend to be highly unstable at best, especially when looking at cross-section slices. Based on our experience as well as results reported by other researchers, better results can be achieved when using a very limited field of view to increase frame rates combined with longitudinal probe positioning so that the shear waves propagate along the muscle fibers. However, in that case, we lose the bigger picture regarding which parts of the muscle are activating, and the potential frame rate is still limited.

\section{Conclusions and Future Work}
\label{sec:conclusion}

In this paper, we have presented a Deep Learning framework for the recovery of sound speed maps from plane wave ultrasound channel data. Results on synthetic data are more than an order of magnitude better than our target accuracy, showing that this framework has great potential for clinical purposes.

Initial real data results are also highly encouraging, although more research is required to improve the results, create calibrated phantoms for validation and improve training as well as develop better simulation techniques to better train the network to deal with real data.

\bibliographystyle{IEEEtran}
\bibliography{IEEEabrv,DeepLearning,DeepLearning2}

\end{document}